\documentclass{article}
\usepackage{hyperref}
\usepackage{graphicx}
\usepackage{multirow}
\usepackage{array}
\usepackage{booktabs}       % professional-quality tables
\usepackage{url}            % simple URL typesetting
\usepackage{algorithmic}
\usepackage{algorithm}
\usepackage{amssymb}
\usepackage{nomencl}
\makenomenclature
\usepackage{lineno,hyperref}

\usepackage{arxiv}

\usepackage[utf8]{inputenc} % allow utf-8 input
\usepackage[T1]{fontenc}    % use 8-bit T1 fonts
\usepackage{hyperref}       % hyperlinks
\usepackage{url}            % simple URL typesetting
\usepackage{booktabs}       % professional-quality tables
\usepackage{amsfonts}       % blackboard math symbols
\usepackage{nicefrac}       % compact symbols for 1/2, etc.
\usepackage{microtype}      % microtypography
\usepackage{lipsum}

\title{DEVDAN: Deep Evolving Denoising Autoencoder \thanks{This paper has been accepted for publication in Neurocomputing 2019. The source code is available in \url{https://www.researchgate.net/publication/335775518_DEVDAN_Code_mFile}.}}

\author{
Andri Ashfahani \thanks{Equal contribution} \\
  School of Computer Science and Engineering\\
  Nanyang Technological University, Singapore \\
  \texttt{andriash001@ntu.edu.sg} \\
  \And
  Mahardhika Pratama $^\dagger$ \\
  School of Computer Science and Engineering\\
  Nanyang Technological University, Singapore \\
  \texttt{mpratama@ntu.edu.sg} \\
  \And
   Edwin Lughofer\\
  Johannes Kepler University Linz, Austria\\
  \texttt{edwin.lughofer@jku.at} \\
   \And
 Yew Soon Ong \\
  School of Computer Science and Engineering\\
  Nanyang Technological University, Singapore \\
  \texttt{asyong@ntu.edu.sg} \\
}

\begin{document}
\maketitle

\begin{abstract}
The Denoising Autoencoder (DAE) enhances the flexibility of data stream method in exploiting unlabeled samples. Nonetheless, the feasibility of DAE for data stream analytic deserves in-depth study because it characterizes a fixed network capacity which cannot adapt to rapidly changing environments. Deep evolving denoising autoencoder (DEVDAN), is proposed in this paper. It features an open structure in the generative phase and the discriminative phase where the hidden units can be automatically added and discarded on the fly. The generative phase refines the predictive performance of discriminative model exploiting unlabeled data. Furthermore, DEVDAN is free of the problem-specific threshold and works fully in the single-pass learning fashion. We show that DEVDAN can find competitive network architecture compared with state-of-the-art methods on the classification task using ten prominent datasets simulated under the prequential test-then-train protocol.
\end{abstract}

% keywords can be removed
\keywords{denoising autoencoder, data streams, incremental learning}

\section{Introduction}
The underlying challenge in the design of Deep Neural Networks (DNNs) is seen in the model selection phase where no commonly accepted methodology exists to configure the structure of DNNs \cite{DeepExpandable,IMM2012}. This issue often forces one to blindly choose the structure of DNNs. DNN model selection has recently attracted intensive research where the goal is to determine an appropriate structure for DNNs with the right complexity for given problems. It is evident that a shallow NN tends to converge much faster than a DNN and handles the small sample size problem better than DNNs. In other words, the size of DNNs strongly depends on the availability of samples. This encompasses the development of pruning  \cite{LearningTheNumber}, regularization \cite{parameterprediction}, parameter prediction \cite{parameterprediction}, etc. Most of which start with an over-complex network followed by a complexity reduction scenario to drop the inactive components of DNNs \cite{distilling}. These approaches, however, do not fully fit to handle streaming data problems because they rely on an iterative parameter learning scenario where the tuning phase is iterated across a number of epochs \cite{GamaDataStream}. Moreover, a fixed structure is considered to be the underlying bottleneck of this model because it does not embrace or is too slow to respond to new training patterns as a result of concept change especially if network parameters have converged to particular points \cite{GamaDataStream}. 

In order to further improve the DNNs' predictive performance with the absence of the true class label, an unsupervised learning step is carried out in the pre-training phase, also known as the generative training phase. In the realm of DNNs, the pre-training phase plays a vital role because it addresses the random initialization problem leading to slow convergence \cite{Bengio_2013}. Of the several approaches for the generative phase, DAE, which adopts the partial destruction of the original input features, is considered the most prominent method because it prevents the learning identity function problem and opens the manifold of the original input dimension. Furthermore, noise injected mechanism of DAE functions as some sort of regularization and rejects low variance direction of input features \cite{learning_dynamic_AE,zeng2018facial}. From the viewpoint of the data stream, the generative phase offers a refinement of the predictive model with the absence of true class label. This case is evident due to the fact that data stream often arrives without labels \cite{GamaDataStream}.

\subsection{Related Work}
The ideas of online DNNs have started to attract research attention \cite{DEEPIOT}. In \cite{Zhou_incrementallearning}, online incremental feature learning is proposed using a denoising autoencoder (DAE) \cite{VincentDAE}. The incremental learning aspect is depicted by its aptitude to handle the addition of new features and the merging of similar features. The structural learning scenario is mainly driven by feature similarity and does not fully operate in the one-pass learning mode. \cite{OnlineDeepLearning} puts forward the hedge backpropagation method  to answer the research question as to how and when a DNN structure should be adapted. This work, however, assumes that an initial structure of DNN exists and is built upon a fixed-capacity network. This property imposes the network's capacity to be user-defined, thus being problem-dependent. To the best of our knowledge, the two approaches are not examined with the prequential test-then-train procedure considering the practical scenario where data streams arrive without labels, thus being impossible to first undertake the training process \cite{GamaDataStream}. Although the dynamic-structured DAE has been proposed in \cite{Zhou_incrementallearning}, it has not exploited  the full advantage of a coupled generative and discriminative process - static discriminative part. A further point, it still relies on hyper-parameters thereby making them an ad-hoc solution.

Several methods have been proposed to increase DNNs' network capacity whenever there is concept change. Progressive Neural Networks (PNN) \cite{progressive} learns $K$ different tasks by introducing new columns while freezing old columns. Its extension is presented with the idea of Dynamically Expandable Network (DEN) \cite{DeepExpandable}. DEN makes use of the selective retraining approach where relevant components of old network structure is brought across a new task and enhanced with the splitting-duplicating strategy. Autonomous Deep Learning (ADL) is proposed in \cite{ADL} offering a fully open structure evolving both network depth and width for data stream learning. It utilizes network significance (NS) formula, which can be computed online, to evolve its network structure. Nonetheless, these three methods still ignore the fact that data streams are received with the absence of true class labels. As a result, they do not benefit from any generative phase which is able to refine the DNNs' predictive performance exploiting unlabeled samples \cite{VincentDAE}.

\subsection{Our Approach}
A deep evolving denoising autoencoder (DEVDAN) for evolving data streams is proposed in this paper. DEVDAN presents an incremental learning approach for DAE which features a fully open and single-pass working principle in both generative and discriminative phase. It is capable of starting its generative learning process from scratch without an initial structure. Its hidden nodes can be automatically generated, pruned and learned on demand and on the fly. Note that this paper considers the most challenging case where one has to grow the network from scratch but the concept is directly applicable in the presence of initial structure. The discriminative model relies on a soft-max layer which produces the end-output of DNN and shares the same trait of the generative phase: online and evolving. 

DEVDAN distinguishes itself from incremental DAE \cite{Zhou_incrementallearning} which still relies on hyper-parameters thereby making them an ad-hoc solution. DEVDAN works by means of estimation of  NS leading to the approximation of bias and variance and is free of user-defined thresholds. A new hidden unit is introduced if the current structure is no longer expressive enough to represent the current data distribution - underfitting whereas an inconsequential unit is pruned in the case of high variance - overfitting. In addition, the evolving trait of DEVDAN is not only limited to the generative phase but also the discriminative phase.

Although the NS formula has been proposed in Autonomous Deep Learning (ADL) \cite{ADL},  DEVDAN applies the elastic learning mechanism in both generative and discriminative phases. The self-adaptive mechanism of generative phase aims to enhance the stability of the learning process due to poor network initialization and to condition the network against possible non-stationary environments - virtual drift handling mechanism. This trait also leads to a reformulation of NS formula to work under encoding-decoding mechanism of DAE. Our numerical study exhibits a clear advantage of DEVDAN over ADL using only a single hidden layer structure. %and numerical consistency across five consecutive runs.

The novelty and contribution of our work are primarily four fold: 1) This paper proposes a novel deep evolving DAE (DEVDAN) for data stream analytic. DEVDAN offers a flexible approach to the automatic construction of extracted features from data streams and operates in the one-pass learning fashion; 2) DEVDAN utilizes the coupled-generative-discriminative-training phases. The structural evolution taking place in both phases makes it possible to adapt to concept change with or without label; 3) The NS formula to govern the network evolution in the generative training phase is derived in this paper; 4) The structural learning mechanism is not an ad-hoc solution and is independent of user defined thresholds.

The advantage of DEVDAN has been thoroughly investigated using ten benchmark problems: Rotated MNIST \cite{rotatedMNIST}, Permuted MNIST \cite{permuttedMNIST}, MNIST \cite{lecun-mnisthandwrittendigit-2010}, Forest Covertype \cite{blackard1999comparative}, SEA \cite{SEA}, Hyperplane \cite{MOA}, Occupancy \cite{candanedo2016accurate}, RFID Localization \cite{rfidL}, KDDCup \cite{KDDCup} and HEPMASS \cite{Baldi2014SearchingFE}. DEVDAN is compared against state-of-the-art methods in data stream methods: ADL \cite{ADL}, pENsemble \cite{pENsemble}, OMB \cite{jung2017online}, Incremental Bagging, Incremental Boosting \cite{IncBaggingBoosting} pENsemble+ \cite{pensembleplus} and LEARN++NSE \cite{Learn++NSE}. DEVDAN is also benchmarked to the existing continual learning methods for deep networks: HAT \cite{kirkpatrick2016overcoming} and PNN \cite{progressive}.

DEVDAN numerical results are produced under \textbf{the prequential test-then-train protocol} - standard evaluation procedure of the data stream method \cite{GamaDataStream} where the windowing approach is applied in evaluating the models' performance. That is, a model is independently examined per data batch and the final numerical results are the average across all data batches. Moreover, a model is supposed to predict the entire data points of an incoming data batch rather than only the next data point. The numerical results are statistically validated using the Wilcoxon to confirm that DEVDAN is significantly different than other algorithms.
%test where the null hypothesis is successfully rejected.
%In general, DEVDAN produces comparable performances to these models and even outperforms them on Rotated MNIST, MNIST, KDDCup, and HEPMASS problems. 
%DEVDAN offers a flexible approach to the automatic construction of robust features from data streams and operates in the one-pass learning fashion.

The remainder of this paper is structured as follows. In Section \ref{problem} the problems are formulated. The DEVDAN algorithm is described in Section \ref{devdans}. The proof of concepts presented in Section \ref{poc} discusses the numerical study in 10 problems, comparison of DEVDAN against state-of-the-arts, ablation study and an additional application in a semi-supervised learning problem. Some concluding remarks are drawn in the last section of this paper.

\subsection{List of Symbols}
The next list describes several symbols that is defined and will be later used within the body of this paper

\begin{tabular}{cl}
     $K$ & the number of timestamps \\
     $B$ & data batches $B = [B_1,B_2,\dots,B_k,\dots,B_K]$ \\
     $T$ & the number of data points in a particular batch $B_k$\\
     $X$ & input \\
     $\widetilde X$ & corrupted input input \\
     $y$ & extracted feature\\
     $z$ & reconstructed input\\
     $C$ & true class labels $C = [C_1,C_2,\dots,C_k,\dots,C_K]$\\
     $\hat C$ & predicted output\\
     $n$ & the input space dimension \\
     $n'$ & the number of corrupted input features \\
     $s(.)$ & sigmoid function\\
     $\Phi(.)$ & probit function\\
     $E[.]$ & expected value\\
     $W$ & encoder weight\\
     $b$ & encoder bias\\
     $c$ & decoder bias\\
     $\Theta$ & output weight\\
     $\eta$ & output bias\\
     $R$ & the number of hidden nodes\\
     $\kappa$ & dynamic confidence level of sigma rule in the growing condition\\
     $\chi$ & dynamic confidence level of sigma rule in the pruning condition
\end{tabular}
%\nomenclature{$c$}{Speed of light in a vacuum inertial frame}
%\nomenclature{$h$}{Planck constant}

\section{Problem Formulation}
\label{problem}
Evolving data streams refer to the continuous arrival of data points in a number of timestamps $K$, $B_k\in [B_1,B_2,...,B_K]$, where $B_k$ may consist of a single data point $B_k=X_{1k}\in\Re^{n}$ or be formed as a data batch of a particular size $B_k=[X_{1k},X_{2k},\dots,X_{tk},...,X_{Tk}]\in\Re^{T\times n}$. $n$ denotes the input space dimension and $T$ stands for the size of the data chunk. The size of data batch often varies and the number of timestamps is unknown in practice. In the realm of real data stream environments, data points come into the picture with the absence of true class labels $C_k\in\Re^{T}$. The labeling process is carried out and is subject to the access of ground truth or expert knowledge \cite{GamaDataStream}. In other words, a delay is expected in consolidating the true class labels. Further, the user may have limited access to the ground truth resulting in less number of labeled data. 

This problem also hampers the suitability of the conventional cross-validation method or the direct train-test partition method as an evaluation protocol of the data stream learner. Hence, the so-called \textbf{prequential test-then-train} procedure is carried out here \cite{datastreamevaluation}. That is, data streams are first used to test the generalization power of a learner before being exploited to perform model's update. The performance of a data stream method is evaluated independently per data batch and the final numerical results are taken from the average of model's performances across all time-stamps. Unlike most algorithms in the literature where only next data point is predicted, a challenging case is considered here where a model is supposed to predict a data batch $B_k$ comprising $T$ data points during the testing phase.

The typical characteristic of the data stream is the presence of concept drift formulated as a change of the joint distribution $P(X_t,C_t)\neq P(X_{t-1},C_{t-1})$ \cite{Gamaconceptdrift}. The concept drift is commonly classified into two types: real and virtual. The real concept drift is more dangerous than the virtual drift because the drift shifts the decision boundary, $P(C_t|X_t)\neq P(C_{t-1}|X_{t-1})$, which deteriorates the network performance. Further, this causes a current model created by previously seen concept $B_{k-1}$ being obsolete. This characteristic is similar to the multi-task learning problem where each data batch $B_k$ is of different tasks. Nevertheless, it differs from the multi-task approaches in which all data batches are to be processed by a single model rather than rely on task-specific classifiers. 

These demands call for an online DNN model which is able to construct its network structure incrementally from scratch in respect to data streams distribution. A further point, the generative training phase can be applied to refine the predictive model in an unsupervised fashion while pending for the operator to annotate the true class label of data samples. The generative training phase should be able to handle the so-called virtual drift, distributional change of the input space, by exploiting unlabeled samples. The virtual drift is interpreted by the change of incoming data distribution $P(X_t) \neq P(X_{t-1})$ \cite{Gamaconceptdrift}. These are \textbf{the underlying motivation} of DEVDAN's algorithmic development.

\section{DEVDAN}
\label{devdans}
In this section, we introduce DEVDAN, our proposed incremental learning approach for DAE.
DEVDAN is constructed under the denoising autoencoder \cite{VincentDAE}, a variant of autoencoder (AE) \cite{Hinton_AE} which aims to retrieve the original input information $X$ from the noise perturbation. The masking noise scenario is chosen here to induce partially destroyed input feature vector $\widetilde X$ by forcing its $n'$ elements to zeros. The number of corrupted input variables $n'$ are randomly destructed in every training observation satisfying the joint distribution $P(\widetilde X,X)$. This mechanism brings DAE a step forward of classical AE since it forces the hidden layer to extract more robust features of the predictive problem minimizing the risk of being an identity function. The identity mapping can also be avoided by AE yet the extracted feature dimension should be less than the input dimension which is inappropriate to be implemented in the evolving network. The reconstruction process is carried out via the encoding-decoding scheme formed with the sigmoid activation function as follows \cite{VincentDAE}:
\begin{equation}
y=f_{(W,b)}=s(\widetilde X W+b)\label{encoder}
\end{equation}
\begin{equation}
z=f_{(W',c)}=s(y W'+c)\label{decoder}
\end{equation}
where $W\in\Re^{n \times R}$ is a weight matrix, $b\in\Re^{R},c\in\Re^{n}$ are respectively the bias of hidden units and the decoding function. $R$ is the number of hidden units. The weight matrix of the decoder is constrained such that $W'$ is the transpose of $W$. That is, DAE has a tied weight \cite{VincentDAE}.
\begin{figure*}[t!]
	\begin{centering}
	\includegraphics[scale=0.4]{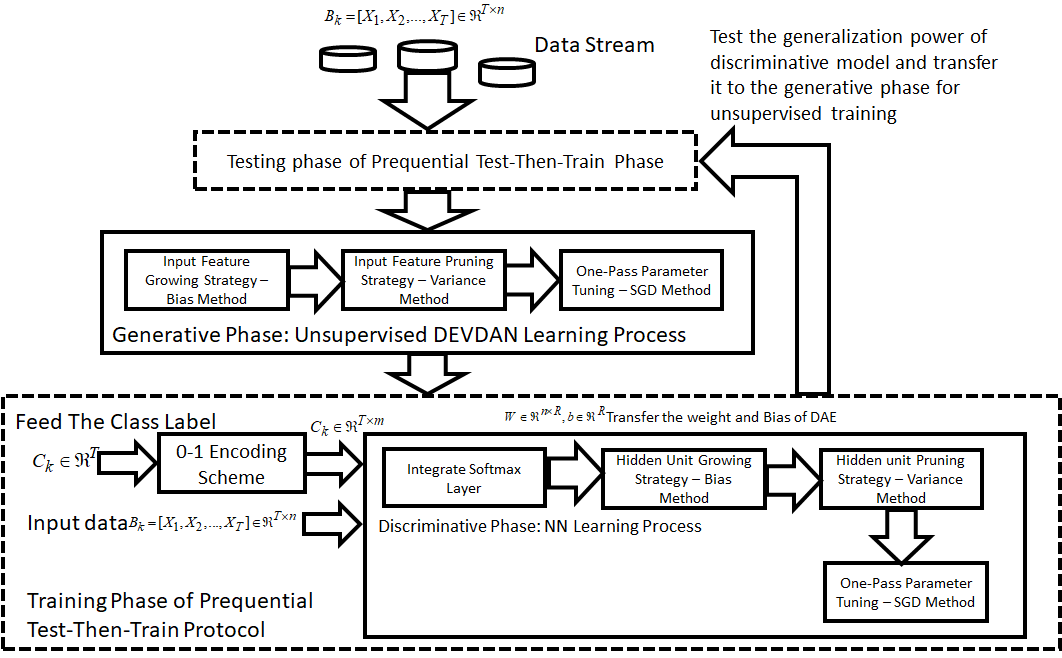}
	\par\end{centering}
	\caption{Learning Mechanism of DEVDAN}
	\label{fig:PHO}
\end{figure*}

%In other words, only a subset of original input features $n-n'$ goes through DAE.
%The typical characteristic of data stream is the presence of concept drift formulated as a change of the joint-class posterior probability $P(Y_t,X_t)\neq P(Y_{t-1},X_{t-1})$ \cite{Gamaconceptdrift}. This situation leads to a current model created by previously induced concept $B_{k-1}$ being obsolete. 
DEVDAN features an open structure where it is capable of initiating its structure from scratch without the presence of a pre-configured structure. Its structure automatically evolves in respect of the NS formula forming an approximation of the network bias and variance. In other words, DEVDAN initially has an extracted input feature where the number of this features incrementally augments $R=R+1$ if it signifies an underfitting situation, high bias, or decreases $R=R-1$ if it suffers from an overfitting situation, high variance. In the realm of concept drift, this is supposed to handle the virtual drift. 
%The parameter tuning scenario is driven by the stochastic gradient descent (SGD) method in \textbf{a single pass} mode \cite{Bengio_Greedy}. 

\begin{algorithm}[!t]
		\caption{Learning policy of DEVDAN}
		\label{devdan}
		\begin{algorithmic}
			\STATE \textit{Initialization}: $R$, $W$, $b$, $c$, $\Theta$, and $\eta$
			\\  \textit{Execute the main loop process}:
			\FOR {$k = 1$ to the number of data batches}
			\STATE \textit{Get}: input data $B_{k}\in\Re^{T\times n}$
			\STATE \textbf{Execute}: Discriminative testing to test DEVDAN's generalization performance
			\STATE \textbf{Execute}: Generative training phase (Algorithm 2) to update the network parameters in an unsupervised manner
			\STATE \textbf{Execute}: Discriminative training (Algorithm 3) to update the network parameters in a supervised manner
			\ENDFOR\\
		\end{algorithmic}
	\end{algorithm}

Once the true class labels of a data batch $B_k$ has been observed $C_k$, the 0-1 encoding scheme is undertaken to construct a labeled data batch $(B_k,C_k)\in\Re^{T\times (n+m)}$  where $m$ stands for the number of the target classes. The discriminative training phase of DEVDAN is carried out once completing the generative training phase. The encoder part is connected to a softmax layer and then trained using the SGD method with momentum in a single pass learning mode. Furthermore, the discriminative training process is also equipped by the hidden unit growing and pruning strategies derived in a similar manner as that of the generative training process. An overview of DEVDAN's learning mechanism is depicted in Fig. \ref{fig:PHO} and DEVDAN's learning procedures are outlined in Algorithm \ref{devdan}, \ref{devdanGen} and \ref{discriminative}. One must bear in mind that DEVDAN's learning scheme can be also applied with an initial network structure.
	
\subsection{Network Significance Formula}
The power of DAE can be examined from its reconstruction error which can be formed in terms of mean square error (MSE) as follows:
\begin{equation}
MSE=\sum_{t=1}^{T}\frac{1}{T}(X_t-z_t)^{2}    
\end{equation}
where $X_t,z_t$ respectively stand for clean input variables and reconstructed input features of DAE. This formula suffers from two bottlenecks for the single-pass learning scenario: 1) it calls for the memory of all data points to understand a complete picture of DAE's reconstruction capability; 2) Notwithstanding that the MSE can be calculated recursively without revisiting preceding samples, this procedure does not examine the reconstruction power of DAE for unseen data samples. In other words, it does not take into account the generalization power of DAE. 
To correct this drawback, let $z$ denotes the estimation of clean input variables $X$ and $E[z]$ stands for the expectation of DAE's output, the NS formula is defined as follows:
\begin{eqnarray}
    NS=\int_{-\infty}^{\infty}(X-z)^{2}p(\widetilde X)d\widetilde X
    \end{eqnarray}
Note that $E[\widetilde X]=\int_{-\infty}^{\infty}\widetilde Xp(\widetilde X)d\widetilde X$ where $p(\widetilde X)$ is the probability density estimation. The NS formula can be defined in terms of the expectation of the squared reconstruction error $E[(X-z)^{2}]$. Several mathematical derivation steps lead to the bias and variance formula as follows:
\begin{eqnarray}
    NS=\int_{-\infty}^{\infty}(X-z)^{2}p(\widetilde X)d\widetilde X\label{nsformula}\nonumber\\
    NS=E[(X-z)^{2}]\nonumber\\
    NS=E[(X-z+E[z]-E[z])^{2}]\nonumber\\
    NS=E[(X-E[z])^{2}]+E[(E[z]-z)^{2}]+2E[(E[z]-z)(X-E[z])]\nonumber\\
    NS=(X-E[z])^{2}+E[(E[z]-z)^{2}]+2E[(E[z]-z)(X-E[z])]\nonumber\\
    NS=(X-E[z])^{2}+E[(E[z]-z)^{2}]\nonumber\\
    NS=(X-E[z])^{2}+E[z^2]-E[z]^2\nonumber\\
    NS=Bias(z)^2+Var(z)\label{eq:NS}
\end{eqnarray}
where the $Bias(z)^2$ and the $Var(z)$ of a random variable $z$ can be expressed as $(X-E[z])^{2}$ and $E[z^2]-E[z]^2$, respectively.

The key for solving (\ref{eq:NS}) is to find the expectation of the recovered input attributes delineating the statistical contribution of DAE. It is worth mentioning that the statistical contribution captures both the network contribution in respect to past training samples and unseen samples. It is thus written as follows:
\begin{equation}
E[z]=\int_{-\infty}^{\infty}s(yW'+c)p(y)dy\label{eq:Ez}
\end{equation}
It is evident that $y$ is induced by the feature extractor $s(\widetilde X W+b)$ and is influenced by partially destroyed input features $\widetilde X$ due to the masking noise. Hence, (\ref{eq:Ez}) is modified as follows:
\begin{eqnarray}
    E[z]=s(E[y]W'+c)\\
    E[y]=\int_{-\infty}^{\infty}s(A)p(A)dA \label{eq:E_y}\\
    A = \widetilde X W+b \nonumber
\end{eqnarray}
Suppose that the normal distribution holds, the probability density function (PDF) $p(A)$ is expressed as $\frac{1}{\sqrt{2\pi(\sigma^{t}_A)^2}}\exp(-\frac{(A - \mu_A^t)^{2}}{2(\sigma^{t}_A)^2})$. It is also known that the sigmoid function $s(A)$ can be approached by the probit function $\Phi(\xi A)$ \cite{Murphy_Machine_Learning} where $\Phi(A)=\int_{-\infty}^A \mathcal{N}(\theta|0,1)d\theta$ and $\xi^{2}=\pi/8$. Following the result of \cite{Murphy_Machine_Learning}, the expectation of $y$, $E[y]$, can be obtained from (\ref{eq:E_y}) as follows:
\begin{eqnarray}
    E[y]=\int_{-\infty}^{\infty}s(A)p(A)dA \thickapprox \int_{-\infty}^{\infty}\Phi(\xi A)p(A)dA\\
    E[y] \thickapprox s(\frac{\mu_A^t}{\sqrt{1+\pi (\sigma^{t}_A)^2/8}})
\end{eqnarray}
where $\mu_A^t$ and $\sigma_A^t$ are respectively the mean and standard deviation of $A$ at the $t-th$ time instant which can be calculated recursively from streaming data $\widetilde X$. The final expression of $E[z]$ is formulated as follows:
\begin{equation}
E[z]=s(E[y]W'+c) \label{eq:E_z2}
\end{equation}
where (\ref{eq:E_z2}) is a function of two sigmoid functions. This result enables us to establish the $Bias(z)^{2}=(X-E[z])^{2}$ in (\ref{eq:NS}).

Let's recall $Var(z)=E[z^{2}]-E[z]^{2}$. The second term $E[z]^{2}$ can be obtained by squaring (\ref{eq:E_z2}) while the first term $E[z^{2}]$ can be written as follows:
\begin{equation}
    E[z^{2}]=s(E[y^{2}]W'+c)
\end{equation}
Due to the fact that $y^{2}=y*y$ , it is obvious that $y^{2}$ is IID variable which allows us to go further as follows:
  \begin{eqnarray}
    E[z^{2}]=s(E[y]E[y]W'+c)\\
    E[z^{2}]=s(E[y]^{2}W'+c) \label{eq:Ez^}
   \end{eqnarray}
Consolidating all the results of (\ref{eq:E_z2}) - (\ref{eq:Ez^}), the final expression of the NS formula is established. $Bias(z)^{2}$ is utilized to control the hidden unit growing, whereas $Var(z)$ are useful to control the hidden pruning. Note that the NS formula of DEVDAN in the generative phase is different from NS formula used in ADL, because it is derived from the expectation of the reconstruction error $(X-z)^2$ instead of the output error $(C_t-\hat{C}_t)^{2}$ and has to accommodate interconnected decoding and encoding part of DAE \cite{ADL}.

The NS formula has been introduced in \cite{ADL} to govern the hidden node evolution of ADL. It is derived from the squared predictive error to detect network performance. In our approach, the NS formula in the generative training phase is derived from the expectation of squared reconstruction error leading to the popular bias and variance formula as per in (\ref{eq:NS}). It examines the quality of DAE by directly inspecting the possible underfitting or overfitting situation and capturing the reliability of an encoder-decoder model across the overall data space given particular data distribution. A high NS value indicates either a high variance problem (overfitting) or a high bias problem (underfitting) which cannot be simply portrayed by a system error index. In other words, this formula helps to find the network architecture satisfying the bias and variance trade-off so that the network achieves low reconstruction error on a given problem. Moreover, the NS formula is computationally inexpensive because it can be calculated recursively and does not require to store previously seen samples. %A further discussion about DEVDAN's complexity is outlined in Subsection 3.3.
%In the realm of Neural Architecture Search (NAS), this approach can be categorized as a performance estimation strategy which is able to estimate the predictive performance on unseen samples \cite{elsken2018neural}. 

\subsection{Generative Training Phase}
This subsection formalizes the generative training phase of DEVDAN.
\begin{algorithm}
	\caption{Generative training phase}
	\label{devdanGen}
	\begin{algorithmic}
    \STATE \textit{Get}: input data $B_{k}\in\Re^{T\times (n+m)}$
		\STATE \textit{Get}: $W$, $b$, and $R$
		\FOR{$t = 1$ to $T$}
		\STATE \textit{Mask}: Original input $X_{tk}$
	\STATE \textit{Execute}: feedforward operation via (\ref{encoder})
\STATE \textit{Calculate}: $e_t = X_t-z_t$,  $\mu_{A}^t$, $\sigma_{A}^t$, $E[z]$, and $E[z^2]$
\STATE \textit{Calculate}: $\mu_{Bias}^t$, $\sigma_{Bias}^t$, $\mu_{Var}^t$, and $\sigma_{Var}^t$ utilizing $E[z]$ and $E[z^2]$
		\STATE \textbf{Hidden node growing mechanism}:
\IF{(\textbf{$\mu_{Bias}^{t} + \sigma_{Bias}^{t}$} $\geq$ $\mu_{Bias}^{min} + \kappa\sigma_{Bias}^{min}$ )}
		\STATE $R = R+1$ 
	\STATE \textit{Initialization}: $W_{new} = -e_t$, $b_{new} = [-1,1]$
	\STATE \textit{Reset}: $\mu_{Bias}^{min}$ and $\sigma_{Bias}^{min}$
		\STATE $grow = 1$
		\ELSE
		\STATE $grow = 0$
		\ENDIF
		\STATE \textbf{Hidden node pruning mechanism}:
\IF{(\textbf{$\mu_{Var}^{t} + \sigma_{Var}^{t}$} $\geq$ $\mu_{Var}^{min} + 2\chi\sigma_{Var}^{min}$ ), (\textbf{$grow$} $=$ $0$ ), and (\textbf{$R$} $>$ $1$ )}
		\FOR {$i=1$ to $R$}
	\STATE \textit{Calculate}: $HS$ via (\ref{eq:NS_expect})
		\ENDFOR
	\STATE \textit{Prune}: hidden node with the smallest HS
		\STATE $R = R-1$
	\STATE \textit{Reset}: $\mu_{Var}^{min}$ and $\sigma_{Var}^{min}$
		\ENDIF
	\STATE \textit{Execute}: backpropagation based on (21)
		\STATE \textit{Update}: $W$, $b$, and $c$
		\ENDFOR
	\end{algorithmic}
\end{algorithm}

\subsubsection{Hidden Unit Growing Strategy}\label{AA}
The hidden unit growing condition is derived from a similar idea to statistical process control which applies the statistical method to monitor the predictive quality of DEVDAN and does not rely on the user-defined parameter \cite{Gama2006,Gamaconceptdrift}. Nevertheless, the hidden node growing condition is not modeled as the binomial distribution here because DEVDAN is more concerned about how to reconstruct corrupted input variables rather than performing binary classification. Because the underlying goal of the hidden node growing process is to relieve the high bias problem, a new hidden node is added if the following condition is satisfied:
\begin{equation}
    \mu_{Bias}^{t} + \sigma_{Bias}^{t} \geq \mu_{Bias}^{min} + \kappa\sigma_{Bias}^{min}\label{eq:HUgrowing}
\end{equation}
where $\mu_{Bias}^{t}$ and $\sigma_{Bias}^{t}$ are respectively the mean and standard deviation of $Bias(z)^2$ at the $t-th$ time instant while $\mu_{Bias}^{min}$ and $\sigma_{Bias}^{min}$ are the minimum mean and the minimum standard deviation of $Bias(z)^2$ up to the $t-th$ observation. These variables are computed with the absence of previous data samples by simply updating their values whenever a new sample becomes available. Moreover, $\mu_{Bias}^{min}$ and $\sigma_{Bias}^{min}$ have to be reset once (\ref{eq:HUgrowing}) is satisfied. This setting is also formalized from the fact that the $Bias(z)^2$ should decrease while the number of training observations increases as long as there is no change in the data distribution. On the other hand, a rise in the $Bias(z)^2$ signals the presence of concept drift which cannot be addressed by simply learning the DAE's parameters.
%Our numerical study also investigates the reset scenario of hidden unit growing mechanism. That is, the numerical results of DEVDAN is analyzed with the use of two different reset scenarios - $\mu_{Bias}^{t}$, $\sigma_{Bias}^{t}$,  $\mu_{Bias}^{min}$, $\sigma_{Bias}^{min}$ are reset (DEVDAN-R) while only $\mu_{Bias}^{min}$, $\sigma_{Bias}^{min}$ are reset in the original DEVDAN algorithm.

The condition (\ref{eq:HUgrowing}) is derived from the so-called sigma rule where $\kappa$ governs the confidence degree of sigma rule. The dynamic constant $\kappa$ is selected as $(1.3\exp(-{Bias(z)^2})+0.7)$ which leads $\kappa$ to revolve around $1$ (in high bias situation) to $2$ (in low bias condition), meaning that it attains the confidence level of 68.2\% to 95.2\%. This strategy aims to improve the flexibility of hidden unit growing process which adapts to the learning context and addresses the problem-specific nature of the static confidence level. A high bias signifies an underfitting situation which can be resolved by adding the complexity of network structure while the addition of hidden unit should be avoided in the case of low bias to prevent the variance increase.  

Once a new hidden node is appended, its parameters, $b$ is randomly sampled from the scope of $[-1,1]$ for simplicity while $W$ is allocated as $-e$. This formulation comes from the fact that a new hidden unit should drive the error toward zero. In other words, $e=X_t-s(y_tW'+c)+s_{R+1}(y_tW_{R+1}^{'}+c)=0$ where $R$ is the number of hidden units or extracted features. New hidden node parameters play a crucial role to assure improvement of reconstruction capability and to drive to a zero reconstruction error. It is accepted that the scope $[-1,1]$ does not always ensure the model's convergence. This issue can be tackled with adaptive scope selection of random parameters \cite{SCN}. 
%Another method commonly used to randomly initialize the network parameters is the Xavier initialization \cite{xavierinitialization}. This method is able to initialize the weights in such a way that the variance remains the same for input and output of a layer. This enables to maintain the back-propagated gradient from exploding to high value or vanishing to zero. At the beginning of the learning process, DEVDAN utilizes this method to create the initial parameters $[W_1,b_1,c_1,\Theta_1,\eta_1]$.

In our numerical study, we also investigate the case where $\mu_{Bias}^{t}$, $ \sigma_{Bias}^{t}$, $\mu_{Bias}^{min}$, and $\sigma_{Bias}^{min}$ are reset (DEVDAN-R) if a new neuron is added. This strategy, however, worsens the training performance. This issue is likely caused by the characteristic of the NS formula measuring the network's generalization power meaning that poor network performance must be seen with respect to previous samples as well. Setting empirical mean and standard deviation to zero during the addition of a new hidden unit causes loss of information. Moreover, resetting $\mu_{Bias}^{min}$ and $\sigma_{Bias}^{min}$ suffices to assign new level in respect to the current data distribution. A similar approach is adopted in the drift detection method \cite{Gama2006}.

\subsubsection{Hidden Unit Pruning Strategy}
The overfitting problem occurs mainly due to a high network variance resulting from an over-complex network structure. The hidden unit pruning strategy helps to find a lower dimensional representation of feature space by discarding its superfluous components. Because a high variance designates the overfitting condition, the hidden unit pruning strategy starts from the evaluation of the model's variance. The same principle as the growing scenario is implemented where the statistical process control method is adopted to detect the high variance problem as follows:
\begin{equation}
        \mu_{Var}^{t} + \sigma_{Var}^{t} \geq \mu_{Var}^{min} + 2\chi\sigma_{Var}^{min}\label{eq:HUpruning}
\end{equation}
where $\mu_{Var}^{t}$ and $\sigma_{Var}^{t}$ respectively stand for the mean and standard deviation of $Var(z)$ at the $t-th$ time instant while $\mu_{Var}^{min}$ and $\sigma_{Var}^{min}$ denote the minimum mean and minimum standard deviation of $Var(z)$ up to the $t-th$ observation. The variable $\chi$, selected as $(1.3\exp(-{Var(z)})+0.7)$, is a dynamic constant controlling the confidence level of the sigma rule. The term 2 is arranged in (\ref{eq:HUpruning}) to overcome a direct-pruning-after-adding problem which may take place right after the feature growing process due to the temporary increase of network variance. The network variance naturally alleviates as more observations are encountered. Note that $Var(z)$ can be calculated with ease by following the mathematical derivation of the NS formula in (\ref{eq:E_z2}) - (\ref{eq:Ez^}). Moreover, $\mu_{Var}^{min}$, $\sigma_{Var}^{min}$ are reset when (\ref{eq:HUpruning}) is satisfied. No reset is applied to $\mu_{Var}^{t},\sigma_{Var}^{t}$ because a high variance case must be judged with respect to previous cases. Note that the pruning scenario is not designed for drift detection.

After (\ref{eq:HUpruning}) is identified, the contribution of each hidden unit is examined. Inconsequential hidden unit is discarded to reduce the overfitting situation. The significance of a hidden unit is tested via the concept of network significance, adapted to evaluate the hidden unit statistical contribution. This method can be derived by checking the hidden node activity in the whole corrupted feature space $\widetilde X$. The significance of the $i-th$ hidden node is defined as its average activation degree for all possible data samples as follows:
\begin{eqnarray}
    HS_i=\lim_{T\to\infty}\sum_{t=1}^{T}\frac{s(A_i)}{T} \label{eq:NS_i2}\\
    A_i = \widetilde X W_i + b_i \nonumber
\end{eqnarray}
where $W_i,b_i$ stand for the connective weight and bias of the $i-th$ hidden node. Suppose that data are sampled from a certain PDF, Eqn. (\ref{eq:NS_i2}) can be derived as follows:
\begin{equation}
    HS_i=\int_{-\infty}^{\infty} s(A_i) p(A_i) d A_i \thickapprox \int_{-\infty}^{\infty}\Phi(\xi A_i)p(A_i)dA_i \label{eq:NS_i3}
\end{equation}
Because the decoder is no longer used and is only used to complete a feature learning scenario, the importance of the hidden units is examined from the encoding function only. As with the growing strategy, (\ref{eq:NS_i3}) can be solved from the fact that the sigmoid function can be approached by the Probit function. The importance of the $i-th$ hidden unit is formalized as follows:
\begin{equation}
    HS_i\thickapprox s(\frac{\mu_{A_i}^t}{\sqrt{1+\pi (\sigma^{t}_{A_i})^2/8}})\label{eq:NS_expect}
\end{equation}
where $\mu_{A_i}^t$ and $\sigma_{A_i}^t$ respectively denote the mean and standard deviation of $A_i$ at the $t-th$ time instant. Because the significance of the hidden node is obtained from the limit integral of the sigmoid function given the normal distribution, (\ref{eq:NS_expect}) can be also interpreted as the expectation of $i-th$ sigmoid encoding function. It is also seen that (\ref{eq:NS_expect}) delineates the statistical contribution of the hidden unit in respect to the recovered input attribute. A small HS value implies that $i-th$ hidden unit plays a small role in recovering the clean input attributes $x$ and thus can be ruled out without significant loss of accuracy.

Since the contribution of $i-th$ hidden unit is formed in terms of the expectation of an activation function, the least contributing hidden unit having the minimum $HS$ is deemed inactive. If the overfitting situation occurs or (\ref{eq:HUpruning}) is satisfied, the pruning process encompasses the hidden unit with the lowest $HS$ as follows:
\begin{equation}
 Pruning \longrightarrow \min_{i=1,...,R} HS_i \label{eq:pruning}
\end{equation}
The condition (\ref{eq:pruning}) aims to mitigate the overfitting situation by getting rid of the least contributing hidden unit. This condition also signals that the original feature representation can be still reconstructed with the rest of $R-1$ hidden units. Moreover, this strategy is supposed to enhance the generalization power of DEVDAN by reducing its variance. 

\subsubsection{Parameter Learning Strategy of Generative Training Phase}
The growing and pruning strategies work alternately with the parameter adjustment mechanism. The network parameters are updated using the SGD method after the structural learning strategy is carried out. Since data points are normalized into the range of $[0,1]$ and are indeed real-valued inputs \cite{Bengio_Greedy}, the SGD procedure is derived using the sum of squared differences loss function as follows:
\begin{equation}
W,b,c=\arg\min_{W,b,c}\sum_{t=1}^{T}\frac{1}{T}L(X_t,z_t)
\end{equation}
\begin{equation}
    L(X_t,z_t)=\frac{1}{2}\sum_{t=1}^{T}(X_t - z_t)^2
\end{equation}
where $X_t\in\Re^{n}$ is the noise-free input vector and $z_t\in\Re^{n}$ is the reconstructed input vector. $T$ is the number of samples observed thus far. The SGD method is utilized in the parameter learning scenario to update $W,b,c$. The first order derivative in the SGD method is calculated with respect to the tied weight constraint $W'=W^{T}$. Note that the parameter adjustment step is carried out under a dynamic network which commences with only a single input feature $R=1$ and grows its network structure on demand.

The generative training phase allows the model's structure to be self-organized in an unsupervised manner. The concept of DAE learns the robust feature by opening the manifold of the learning problem. The information learned in this phase can be utilized to perform better in the discriminative training phase. The reason is that some features that are meaningful for the generative phase may also be meaningful for the discriminative phase. Furthermore, DEVDAN addresses the random initialization problem by implementing the generative training phase. This training phase helps to move the network parameters into inaccessible region \cite{goodfellow2016deep}. As a result, this expedites parameter's convergence in the discriminative training phase. All of which can be committed while pending for operator to feed the true class labels $C_k$. Although DEVDAN is realized in the single hidden layer architecture, it is modifiable to the deep structure with ease by applying the greedy layer-wise learning process \cite{Bengio_Greedy}. 
%The concept of DAE discovers the salient structure of input space by opening manifold of learning problem.

\subsection{Discriminative Training Phase}
Once the true class labels $C_k=[C_{1k},C_{2k},\dots,C_{Tk}]\in\Re^{T}$ are obtained, the 0-1 encoding scheme is applied to craft the target vector $C_k\in\Re^{T\times m}$ where $m$ is the number of the target class. That is, $C_o=1$ if only if a data sample $X_t$ falls into $o$-th class. A generative model is passed to the discriminative training phase added with a softmax layer to infer the final classification decision as follows:
\begin{equation}
    \hat{C_t}=softmax(s(X W+b) \Theta+\eta)\label{forwarddisc}
\end{equation}
where $\Theta\in\Re^{R \times m}$ and $\eta\in\Re^{m}$ denote the output weight vector and bias of discriminative network respectively while the softmax layer outputs probability distribution across $m$ target classes. The parameters, $W,b,\Theta,\eta$ are further adjusted using the labeled data chunk $(B_k,C_k)\in\Re^{T\times(n+m)}$ via the SGD method with momentum using only a single epoch. The optimization problem is formulated as follows:
\begin{equation}
W,b,\Theta,\eta = \arg\min_{W,b,\Theta,\eta}\sum_{t=1}^{T}\frac{1}{T}L(C_t,\hat{C}_t) \label{backdisc}
\end{equation}
where $L(C_t,\hat{C}_t)$ is the cross-entropy loss function. The adjustment process is executed in the one-pass learning fashion and per-sample adaptation process.
%The learning rate and the momentum coefficient are 0.01 and 0.95, respectively.

The structural learning scenario also occurs in the discriminative training phase where the NS method can be formulated in respect to the squared predictive error rather than reconstruction error. The similar derivation is applied here yet the difference only exists in the output expression of the discriminative model as $s(X_t W+b)\Theta+\eta$ instead of the encoding and decoding scheme as shown in Eqns. (\ref{encoder}), (\ref{decoder}). It should be noted that in discriminative training phase $\mu_A^t$ and $\sigma_A^t$ of $E[y]$ are calculated using clean input $X$ instead of $\widetilde X$. Finally, the $Bias^2(\hat{C})$ and $Var(\hat{C})$ are formalized as $(E[\hat{C}]-C)^2$ and $(E[\hat{C}^2]-E[\hat{C}]^2)$, respectively. The hidden node growing and pruning conditions still refer to the same criteria (\ref{eq:HUgrowing}), (\ref{eq:HUpruning}) yet the new weight and bias are initialized using Xavier initialization \cite{ADL,xavierinitialization}. 
%This method is able to initialize the weights in such a way that the variance remains the same for input and output of a layer. This enables to maintain the back-propagated gradient from exploding to high value or vanishing to zero. At the beginning of the learning process, DEVDAN utilizes this method to create the initial parameters $[W_1,b_1,c_1,\Theta_1,\eta_1]$.
\begin{figure}[!t]
	    \begin{centering}
	        \includegraphics[scale=0.5]{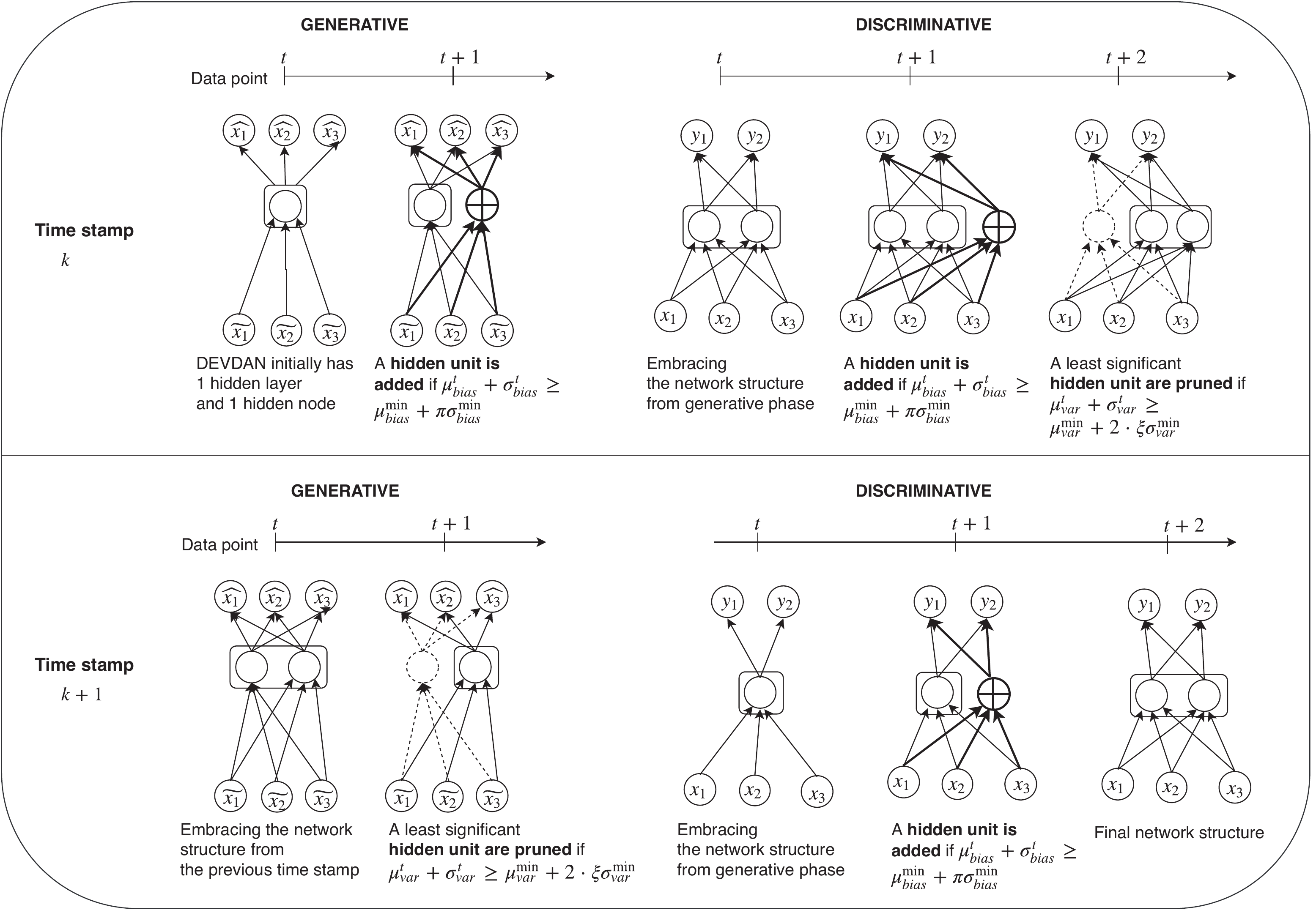}
	    \par\end{centering}
	    \caption{The hidden unit evolution of DEVDAN algorithm. It starts the learning process from scratch with a single hidden unit. It can evolve the network structure both in the generative and discriminative phase if the growing or pruning condition is satisfied. At the end time stamp $k+1$, it has 2 hidden units.}
	    \label{fig:PHO3}
\end{figure}

%explain avoid RI. This aims to learn the robust features and condition .  the absence of true class label $C_{k+1}$.  exploiting unlabeled samples
%consists of generative training phase which works in an unsupervised manner. By implementing this training phase, DEVDAN specifically circumvents the random initialization problem as it is able to condition the network parameters into the region that they do not escape \cite{goodfellow2016deep}. When the data streams $B_{k+1}$ arrive without label, the generative phase updates the model of the previous time stamp. After that, the discriminative phase further improves the performance once the operator has completed the labeling process. This creates a truly continual learning cycle. The coupled-generative-discriminative-training phases helps DEVDAN to handle a semi-supervised learning problem where the access to the ground truth is limited. The evolution of DEVDAN's network structure is illustrated in Fig. \ref{fig:PHO3}.
In the discriminative phase, DEVDAN learning strategy is similar to ADL \cite{ADL} yet one must bear in mind that DEVDAN implements the coupled-generative-discriminative-training phases which work in both unsupervised and supervised manner. The generative phase trains the network exploiting unlabeled samples and specifically circumvents the random initialization problem as it is able to condition the network parameters into the region that they do not escape \cite{goodfellow2016deep}. After that, the discriminative phase further improves the performance once the operator has completed the labeling process. This creates a truly continual learning cycle. A further point, the coupled-generative-discriminative-training phases help DEVDAN to handle a semi-supervised learning problem where the numerical results are discussed in Subsection \ref{latgt}. The evolution of DEVDAN's network structure is illustrated in Fig. \ref{fig:PHO3}.
\begin{algorithm}
		\caption{Discriminative training phase}
		\label{discriminative}
		\begin{algorithmic}
			\STATE \textit{Define}: input-output pair $(B_k,C_k)\in\Re^{T\times (n+m)}$
			\STATE \textit{Get}: $W$, $b$, and $R$
			\FOR {$t = 1$ to $T$}
			\STATE \textit{Execute}: feedforward operation via (\ref{forwarddisc})
			\STATE \textit{Calculate}: $e_t = C_t-\hat{C}_t$,  $\mu_{A}^t$, $\sigma_{A}^t$, $E[\hat{C}_t]$, and $E[\hat{C}_t^2]$
			\STATE \textit{Calculate}: $\mu_{Bias}^t$, $\sigma_{Bias}^t$, $\mu_{Var}^t$, and $\sigma_{Var}^t$ utilizing $E[\hat{C}_t]$ and $E[\hat{C}_t^2]$
			\STATE \textbf{Hidden node growing mechanism}:
			\IF{(\textbf{$\mu_{Bias}^{t} + \sigma_{Bias}^{t}$} $\geq$ $\mu_{Bias}^{min} + \kappa\sigma_{Bias}^{min}$ )}
			\STATE $R = R+1$ 
			\STATE \textit{Initialization}: $W_{new}$, $\Theta_{new}$, and $b_{new}$
			\STATE \textit{Reset}: $\mu_{Bias}^{min}$ and $\sigma_{Bias}^{min}$
			\STATE $grow = 1$
			\ELSE
			\STATE $grow = 0$
			\ENDIF
			\STATE \textbf{Hidden node pruning mechanism}:
			\IF{(\textbf{$\mu_{Var}^{t} + \sigma_{Var}^{t}$} $\geq$ $\mu_{Var}^{min} + 2\chi\sigma_{Var}^{min}$ ), (\textbf{$grow$} $=$ $0$ ), and (\textbf{$R$} $>$ $1$ )}
			\FOR {$i=1$ to $R$}
			\STATE \textit{Calculate}: $HS$ via (\ref{eq:NS_expect})
			\ENDFOR
			\STATE \textit{Prune}: hidden node with the smallest HS
			\STATE $R = R-1$
			\STATE \textit{Reset}: $\mu_{Var}^{min}$ and $\sigma_{Var}^{min}$
			\ENDIF
			\STATE \textit{Execute}: backpropagation based on (\ref{backdisc})
			\STATE \textit{Update}: $W$, $b$, $\Theta$, and $\eta$
			\ENDFOR\\
		\end{algorithmic}
	\end{algorithm}

Note that data stream always comes into picture with the absence of true class labels in practice. Our experiment reflects those facts as a result of the prequential test-then-train procedure. Further, we may arrive at the situation where we have limited access to the ground truth. Consequently, the number of labeled data can be less than the number of unlabeled data. In order to examine DEVDAN's performance in this situation, we have conducted an additional experiment simulating the real-world case where there exists a portion of unlabeled data in every data batch $B_k$. %The pseudocode of DEVDAN's generative and discriminative phases are outlined in the Algorithm \ref{devdan}, \ref{devdanGen} and \ref{discriminative}.

\subsection{Complexity Analysis}
Using the notation in this paper, Table \ref{result-2-1-1-1} presents a summary of the worst scenario of time complexity of the above training phases for a single data sample. It can be seen that the computational cost of DEVDAN lies in the parameter adjustment mechanism, especially when the number of hidden units and input dimension are very large. For a training data stream comprising $T$ samples and $K$ batches, the total time complexity of the learning process is given by (\ref{timecom}): 
\begin{eqnarray}
    O([12\times[R\times n]+4\times[R\times m]+4]\times T \times K)\label{timecom}
\end{eqnarray}
Such complexity is fairly low as DEVDAN's time complexity has no quadratic-time complexity $O(n^2)$. On the other hand, the overall space complexity of the learning procedure is given by (\ref{spacecom}):
\begin{equation}
    O([6\times[R\times n]+3\times[R\times m]+8])\label{spacecom}
\end{equation}
This storage requirement, which can be largely attributed to the size of the gradients and the weights, is reasonable.

These facts suggest that DEVDAN is scalable and, at the same time, able to cope with a fast data stream environment. This benefit is evident in our numerical studies where DEVDAN's training time is faster than those Incremental Bagging, pENsemble, pENsemble+ and LEARN++NSE (see Table \ref{result}).
%has no quadratic-time complexity $O(n^2)$. 

\begin{table}[!t]
\caption{Time complexity of the DEVDAN training phases}
\centering{}\label{result-2-1-1-1} \scalebox{0.8}{ %
\begin{tabular}{llrr}
\toprule
Training phase &  & Time complexity  & Description\tabularnewline
\hline 
Generative & Hidden unit growing  & $O(2\times[R\times n]+1)$ & Calculate the growing condition and create a new node.\tabularnewline
 & Hidden unit pruning & $O(2\times[R\times n]+1)$ & Calculate the pruning condition and delete a new node.\tabularnewline
 & SGD & $O(2\times[R\times n])$ & Calculate the gradient of parameters.\tabularnewline
 & Parameter adjustment & $O(2\times[R\times n])$ & Apply the gradient to update the parameters.\tabularnewline
\hline 
Discriminative & Hidden unit growing  & $O([R\times n]+[R\times m]+1)$ & Calculate the growing condition and create a new node.\tabularnewline
 & Hidden unit pruning & $O([R\times n]+[R\times m]+1)$ & Calculate the pruning condition and delete a new node.\tabularnewline
 & SGD & $O([R\times n]+[R\times m])$ & Calculate the gradient of parameters.\tabularnewline
 & Parameter adjustment & $O([R\times n]+[R\times m])$ & Apply the gradient to update the parameters.\tabularnewline
\bottomrule
\end{tabular}} 
\end{table}

\section{Proof of Concepts}
\label{poc}
To test the effectiveness of DEVDAN, we apply it to standard supervised learning benchmarks, conduct the statistical test to confirm the significance of DEVDAN's performances and provide an extensive ablation study to measure the contribution of each of DEVDAN’s components. As an additional application, we consider a real-world problem where we have limited-access-to-the-ground-truth (Subsection \ref{latgt}). In this experiment, the portion of labeled data in each data batch varies from 25\%, 50\% and 75\%.
%we only have access to the portion of labeled data in each data batch, e.g. 25\%, 50\% and 75\%.

\subsection{Implementation Details} \label{implementationdetails}
In all experiments, DEVDAN starts the learning process from scratch by having a hidden unit. The evolving mechanism of DEVDAN is free of user-defined threshold. We utilize SGD method to adjust parameters and use learning rates of 0.01 and 0.001 for discriminative and generative phases, respectively. In the discriminative phase, we use a momentum coefficient of 0.95. Small learning rates are preferred to make the training process more stable, whereas a high momentum coefficient to reduce the risk of being entrapped in local minima. We use 10\% masking noise to get $\widetilde{X}$ from the original input $X$. Note that these values is fixed in all experiments to demonstrate that DEVDAN is not an ad-hoc method. The parameter adjustment mechanism is executed in a single-pass manner to simulate the most difficult situation in continual learning and to demonstrate that the NS formula can be calculated in one-pass learning fashion. DEVDAN is executed in 5 consecutive runs during the simulation and the numerical results of  the lowest classification rate are reported in Table \ref{result}.
%The results of all runs are tabulated in Table \ref{runs}.
%The SGD with momentum method is employed to adjust the network parameter. The learning rates for the discriminative and generative training phases are selected as 0.01 and 0.001, respectively. 
%The network parameters are adjusted using fixed learning rates, that is selected as 0.01 and 0.001 for discriminative and generative phase, respectively. Small learning rates are preferred to make the training process more stable. In the discriminative phase, we implement SGD with momentum to adjust the parameters and we use the fixed momentum coefficient 0.95. A high momentum coefficient is selected because it can reduce the risk of being entrapped in local minima.

The prequential test-then train procedure is followed as our evaluation protocol to simulate real data stream environments. The windowing approach is adopted in the numerical evaluation where the learning performance is regularly evaluated per data batch to forget the effect of past data batches and to better evaluate the model's performance under concept drift \cite{datastreamevaluation}. The final numerical results are the average of numerical results per data batch. It is worth noting that an algorithm here not only produces one-step-ahead prediction but also performs classification of all data points in the data batch during the testing phase. All consolidated algorithms are executed in the same computational platform under MATLAB environments with the Intel(R) Xeon(R) CPU E5-1650 @3.20 GHz processor and 16 GB RAM. The source code of DEVDAN is publicly available, it can be accessed in \url{https://bit.ly/2Jk3Pzf}. We also provide a short video which demonstrates DEVDAN's learning performance.

\subsection{Baseline Algorithms}
The numerical results of DEVDAN are compared against state-of-the-art data stream and continual learning algorithms: ADL \cite{ADL}, HAT \cite{kirkpatrick2016overcoming}, PNN \cite{progressive}, OMB \cite{jung2017online}, pENsemble \cite{pENsemble}, pENsemble+ \cite{pensembleplus}, Incremental Bagging, Incremental Boosting \cite{IncBaggingBoosting} and LEARN++NSE \cite{Learn++NSE}. ADL, PNN, pENsemble and pENsemble+ are able to evolve their network structure on demands, whereas OMB, Incremental Bagging, Incremental Boosting, and LEARN++NSE utilize several learners to execute a classification task. HAT is a prominent continual learning algorithm which is able to preserve previous tasks' information without affecting the current task's learning. We reimplemented each of these algorithms in the same simulation scenario and computational environment to ensure fair comparison. We re-tuned the hyperparameters for each baseline algorithm, which generally resulted in better performance, thereby providing a more competitive experimental setting for testing out DEVDAN.

The learning performance of the consolidated algorithms is evaluated according to six criteria: classification rate (CR), number of parameters (NoP), training time (TrT), testing time (TsT), number of hidden units (HN) and number of hidden layers (HL). The numerical results of pENsemble, pENsemble+, Incremental Bagging,  Incremental Boosting, and LEARN++NSE are not reported in several problems. This is because they are not scalable to face high-dimensional data such as image data. HAT's, PNN's and OMB's execution times are not comparable because it is developed under Python's environments.

\subsection{Dataset Description}
The learning performance of DEVDAN is numerically validated using ten real-world and synthetic data stream problems. This subsection outlines the characteristics of those datasets. At least six of ten problems characterize \textbf{non-stationary properties}, while the remaining four problems feature salient characteristics in examining the performance of the data stream algorithms: big size, high input dimension, etc. The properties of the dataset are outlined on Table \ref{properties}. The ten datasets are detailed as follows:

\textit{Rotated MNIST} \cite{rotatedMNIST}: It forms an extension of the traditional MNIST problem via rotation of original samples \cite{lecun-mnisthandwrittendigit-2010} inducing abrupt concept drifts. That is, the handwritten digits are rotated to arbitrary angles of the $-\pi$ to $\pi$ range, thus inducing the covariate drift.  

\textit{Permuted MNIST} \cite{permuttedMNIST}: This is a modification of the MNIST problem \cite{lecun-mnisthandwrittendigit-2010} which applies several permutations of pixels and features uncorrelated distribution of input samples across each task. In other words, the real drift \cite{GamaDataStream} is present in this dataset. Three permutations are applied in the original MNIST problem resulting in abrupt and recurring drifts. That is, the drift eventually returns to its original concept.  

\textit{MNIST}: This is a popular benchmark problem whose objective is to perform handwritten digit recognition with 10 classes \cite{lecun-mnisthandwrittendigit-2010}. It consists of 70 K data points formed as black and white 28-by-28-pixel images. 

\textit{Forest Covertype}: This data contains information about Forest Covertype from cartographic variables. The classification task is to predict the actual Forest Covertype whose the ground truth was determined from US Forest Service (USFS) Region 2 Resource Information System (RIS) data. This data also contains binary (0 or 1) inputs representing the qualitative independent variables, such as wilderness areas and soil types \cite{blackard1999comparative}. This data contains covariate drift as the input distribution is changing over time.  %Several independent variables were obtained from  US Geological Survey (USGS) and USFS data.

\textit{SEA Problem}: the SEA problem is one of the most popular non-stationary data stream problems in the literature \cite{SEA} which features a binary classification problem formed by the following inequality $f_1+f_2<\theta$ indicating a class $1$ whereas the opposite condition leads to a class $2$. The concept drift is induced by changing the class threshold three times $\theta=4\longrightarrow 7\longrightarrow4\longrightarrow7$ which leads to two drift types: abrupt and recurring. This problem consists of three input attributes in which the third input feature functions as a noise. This problem consists of 100 K data samples and the prequential test-then-train process is simulated with 100 timestamps. Although the SEA problem is a synthetic dataset, the use of a synthetic dataset is important to develop a controlled simulation environment where the type of drift and the time instant when the concept drift occurs is fully deterministic.
%Data points are sampled from the range of $[0,10]$. Moreover, a modified version of SEA problem \cite{Learn++NSE} is put forward here and carries a class imbalanced problem with 5\% to 25\% class proportion. 

\textit{Hyperplane Problem}: the Hyperplane problem characterizes an artificial binary classification problem where the underlying objective is to separate data points into two classes in respect to the position of $d$-dimensional random hyperplane $\sum_{j=1}^{d}w_jx_j>w_o$. The hyperplane problem is taken from the massive online analysis (MOA) - a popular framework in the data stream field \cite{MOA}. This problem puts forward the gradual drift circumstance where data samples are initially drawn from one distribution with a probability of one where this probability gradually weakens up to a point where the second distribution completely replaces the first one. This problem consists of 120 K data samples and is generated with 120 timestamps.

\textit{Occupancy Problem}: This is a real-world multi-variate time series on room occupancy as per the environmental condition of the room. The data set contains 20560 instances, 7 attributes and 2 classes. The true class label of occupancy was derived from time stamped pictures taken every minute \cite{candanedo2016accurate}. There exists covariate drift in this problem. That is a change in the distribution of the input as a result of environmental change over time.

\textit{Indoor RFID Localization Problem}: the indoor RFID localization problem presents a multi-class classification problem which identifies the object's location in the manufacturing shopfloor. RFID reader is placed in different locations and creates four zones in the manufacturing shopfloor leading to a four classes classification problem. The RFID localization problem is undertaken using three input attributes and comprises 281.3 K data samples \cite{rfidL}.
\begin{table}[t]
\caption{Properties of the dataset.}
\begin{centering}
\label{properties} \scalebox{0.8}{ %
\begin{tabular}{lccccc}
\toprule
Dataset  & IA  & C  & DP  & Tasks  & Characteristics\tabularnewline
\hline 
Rotated MNIST  & 784  & 10  & 65K  & 65  & Non-stationary\tabularnewline
Permuted MNIST  & 784  & 10  & 70K  & 70  & Non-stationary\tabularnewline
MNIST  & 784  & 10  & 70K  & 70  & Stationary\tabularnewline
Forest Covertype & 54 & 7 & 581K & 581 & Non-stationary\tabularnewline
SEA  & 3  & 2  & 100K  & 100  & Non-stationary\tabularnewline
Hyperplane  & 4  & 2  & 120K  & 120  & Non-stationary\tabularnewline
Occupancy & 7 & 2 & 20K & 20 & Non-stationary\tabularnewline
RFID Localization & 3 & 4 & 280K & 280 & Stationary\tabularnewline
KDDCup 10\%  & 41  & 2  & 500K  & 500  & Non-stationary\tabularnewline
HEPMASS 19\% & 27 & 2 & 2M & 2000 & Stationary\tabularnewline
\bottomrule
\end{tabular}} 
\par\end{centering}
\centering{}IA: input attributes, C: classes, DP: data points 
\end{table}

\textit{KDDCup Problem}: this dataset presents a network intrusion detection problem formulated as a binary classification problem recognizing attack of network connection \cite{KDDCup}. This problem possesses non-stationary components since it presents various types of intrusions simulated in a military network environment. It was used in the Third International Knowledge Discovery and Data Mining Tools Competition taking place during the KDD-99. Moreover, the KDDcup problem characterizes a high input dimension with $41$ input attributes. In total, there exist 5 M pairs of data samples in the KDD cup problem and only 10\% of which are collected for our numerical study. Five hundred timestamps are set in the prequential test-then-train procedure of our numerical study.

\textit{HEPMASS Problem}: this problem describes the high-energy physic experiments to discover the signatures of exotic particles with unknown mass carried out under the Monte-Carlo simulations \cite{Baldi2014SearchingFE}. The classification task is to separate particle-producing collisions from a background source. This problem consists of 27 input attributes - 22 low-level features and 5 high-level features. A total of more than 10 million samples are generated and only 2 million samples are utilized in our study. %The prequential test-then-train procedure is executed with 2 K timestamps.

\subsection{Results}
Numerical results are summarized in Table \ref{result} and the algorithm's ranks based on their classification performance are presented in Table \ref{ranking}. The trace of bias and variance in the generative and discriminative phases, hidden units, classification rates, loss functions and hidden units per timestamps are portrayed in the Fig. \ref{fig:evolution} and \ref{fig:evolution2}.
\begin{table*}
\caption{The classification rate ranking of consolidated algorithms in ten
problems.}
\centering{}\label{ranking} \scalebox{0.7}{ %
\begin{tabular}{lcccccccccc}
\toprule
 & \multicolumn{10}{c}{Problems}\tabularnewline
\cline{2-11} 
 & R. MNIST  & P. MNIST & MNIST & F. Covertype & SEA  & Hyperplane & Occupancy & RFID & KDDCup $10\%$  & HEPMASS 19\%\tabularnewline
\hline
DEVDAN  & 1  & 2 & 1 & 4 & 5 & 3 & 3 & 5 & 1  & 1\tabularnewline
ADL  & 2 & 1 & 2 & 5 & 2 & 1 & 6 & 4 & 2 & 2\tabularnewline
HAT & 3 & 3 & 3 & 7 & 9 & 9 & 8 & 6 & 3 & 8\tabularnewline
PNN & 4 & 4 & 4 & 8 & 8 & 7 & 9 & 7 & 6 & 9\tabularnewline
OMB & 5 & 5 & 5 & 6 & 6 & 6 & 1 & 2 & 9 & 6\tabularnewline
$Inc_{Boosting}$  & N/A & N/A & N/A & N/A & 10 & 10 & 10 & N/A & 7 & 5\tabularnewline
$Inc_{Bagging}$  & N/A & N/A & N/A & 1 & 7 & 8 & 7 & 1 & 4 & 7\tabularnewline
pENsemble  & N/A & N/A & N/A & 2 & 4 & 2 & 5 & 9 & 5 & 3\tabularnewline
pENsemble+ & N/A & N/A & N/A & 3 & 1 & 5 & 4 & 8 & 8 & 4\tabularnewline
LEARN++.NSE & N/A & N/A & N/A & N/A & 3 & 4 & 2 & 3 & N/A & N/A\tabularnewline
\bottomrule
\end{tabular}} 
\end{table*}

\begin{table}[!t]
\caption{Numerical results of consolidated algorithms.}
\begin{centering}
\label{result} \scalebox{0.6}{ %
\begin{tabular}{ll|r|rrrrrrrrr}
\toprule
 &  & \textbf{DEVDAN}  & ADL  & HAT & PNN & OMB & I. Boosting & I. Bagging & pENsemble & pENsemble+ & L++.NSE\tabularnewline
\hline
\multirow{6}{*}{\rotatebox{90}{Rotated MNIST}} & CR  & \textbf{$\textbf{76.48}\pm\textbf{9.7}$}  & $73.97\pm{9.92}$  & $65\pm{12}$  & $57\pm{13.9}$  & $26\pm{6}$  & N/A & N/A & N/A & N/A & N/A\tabularnewline
 & TrT  & $1.6\pm{0.13}$  & $0.38\pm{0.41}$  & N/A & N/A & N/A & N/A & N/A & N/A & N/A & N/A\tabularnewline
 & TsT  & $0.006\pm{0.001}$  & $0.06\pm{0.02}$  & N/A & N/A & N/A & N/A & N/A & N/A & N/A & N/A\tabularnewline
 & HN  & $48.7\pm{9}$  & $66.98\pm{9.3}$  & $60$  & $750$  & N/A & N/A & N/A & N/A & N/A & N/A\tabularnewline
 & HL  & $1$  & $1.14\pm{0.4}$  & $2$  & $3$  & $3$ & N/A & N/A & N/A & N/A & N/A\tabularnewline
 & NoP  & $(38\pm{8})$K  & $(18\pm{7.5})$K  & $24.9$K  & $530$K  & N/A & N/A & N/A & N/A & N/A & N/A\tabularnewline
\hline 
\multirow{6}{*}{\rotatebox{90}{Permuted MNIST}} & CR  & $76.67\pm{14}$  & \textbf{$\textbf{79.8}\pm\textbf{14.6}$}  & $66\pm{16}$  & $65\pm{13.9}$  & $11\pm{6}$  & N/A & N/A & N/A & N/A & N/A\tabularnewline
 & TrT  & $1.65\pm{0.1}$  & $0.37\pm{0.02}$  & N/A & N/A & N/A & N/A & N/A & N/A & N/A & N/A\tabularnewline
 & TsT  & $0.007\pm{0.001}$  & $0.06\pm{0.001}$  & N/A & N/A & N/A & N/A & N/A & N/A & N/A & N/A\tabularnewline
 & HN  & $67.8\pm{16.8}$  & $20\pm{5}$  & $60$  & $750$  & N/A & N/A & N/A & N/A & N/A & N/A\tabularnewline
 & HL  & $1$  & $1$  & $2$  & $3$  & $3$ & N/A & N/A & N/A & N/A & N/A\tabularnewline
 & NoP  & $(53\pm{14})$K  & $(16\pm{4})$K  & $24.9$K  & $530$K  & N/A & N/A & N/A & N/A & N/A & N/A\tabularnewline
\hline 
\multirow{6}{*}{\rotatebox{90}{MNIST}} & CR  & \textbf{$\textbf{86.12}\pm\textbf{7.8}$}  & $86.07\pm8.22$  & $78\pm{12}$  & $68\pm{13.4}$  & $29\pm{5}$  & N/A & N/A & N/A & N/A & N/A\tabularnewline
 & TrT  & $2.1\pm{0.11}$  & $0.87\pm{0.23}$  & N/A & N/A & N/A & N/A & N/A & N/A & N/A & N/A\tabularnewline
 & TsT  & $0.009\pm{0.002}$  & $0.12\pm{0.05}$  & N/A & N/A & N/A & N/A & N/A & N/A & N/A & N/A\tabularnewline
 & HN  & $68.92\pm{14.9}$  & $108\pm{6.2}$  & $60$  & $750$  & N/A & N/A & N/A & N/A & N/A & N/A\tabularnewline
 & HL  & $1$  & $1.2\pm{0.4}$  & $2$  & $3$  & $2$ & N/A & N/A & N/A & N/A & N/A\tabularnewline
 & NoP  & $(54\pm{13})$K  & $(17\pm{5})$K  & $24.9$K  & $530$K  & N/A & N/A & N/A & N/A & N/A & N/A\tabularnewline
\hline 
\multirow{6}{*}{\rotatebox{90}{Forest Covertype}} & CR  & $83.29\pm9.06$  & $81.97\pm22.47$  & $67\pm{14}$  & $61\pm{8}$  & $71\pm{4}$  & N/A & \textbf{$\textbf{89.86}\pm\textbf{8.62}$}  & \textbf{$83.69\pm{8.57}$}  & $83.31\pm{8.9}$  & N/A\tabularnewline
 & TrT  & $0.6\pm{0.04}$  & $0.17\pm{0.01}$  & N/A & N/A & N/A & N/A & $3.35\pm{1.78}$  & $19.7\pm{1.2}$  & $15.67\pm{4.85}$  & N/A\tabularnewline
 & TsT  & $0.005\pm{0.002}$  & $0.02\pm{0.003}$  & N/A & N/A & N/A & N/A & $5.17\pm{3.24}$  & $0.44\pm{0.03}$  & $0.46\pm{0.03}$  & N/A\tabularnewline
 & HN  & $70.78\pm{19.5}$  & $20\pm{10}$  & $60$  & $60$  & N/A & N/A & $100$  & $1$  & $1$  & N/A\tabularnewline
 & HL  & $1$  & $1$  & $2$  & $3$  & $2$ & N/A & N/A & $1$  & $1.002\pm{0.03}$  & N/A\tabularnewline
 & NoP  & $(4.4\pm{1.2})$K  & $159\pm{81}$  & $2.9$K  & $2.5$K  & N/A & N/A & N/A & $27$  & $27$  & N/A\tabularnewline
\hline 
\multirow{6}{*}{\rotatebox{90}{SEA}} & CR  & $91.12\pm{7.11}$  & $92\pm{6.49}$  & $75\pm{10}$  & $83\pm{6}$  & $88\pm{4}$  & $79.6\pm{6.18}$  & $87.3\pm{10.2}$  & $91.61\pm{5.6}$  & \textbf{$\textbf{92}\pm\textbf{6}$}  & $91.93\pm{5.9}$ \tabularnewline
 & TrT  & $0.49\pm{0.04}$  & $0.16\pm{0.01}$  & N/A & N/A & N/A & $0.0004\pm{0.0002}$  & $1.35\pm{0.79}$  & $0.92\pm{0.09}$  & $0.5\pm{0.1}$  & $2.77\pm{1.61}$ \tabularnewline
 & TsT  & $0.003\pm{0.006}$  & $0.02\pm{0.002}$  & N/A & N/A & N/A & $0.0024\pm{0.0014}$  & $3.63\pm{2.39}$  & $0.45\pm{0.05}$  & $0.3\pm{0.04}$  & $1.39\pm{0.81}$ \tabularnewline
 & HN  & $23.7\pm{7.2}$  & $21\pm{4}$  & $10$  & $33$  & N/A & N/A & $100$  & $2$  & $2.51\pm{0.81}$  & $10$ \tabularnewline
 & HL  & $1$  & $1.01\pm{0.1}$  & $2$  & $3$  & $2$ & N/A & N/A & $1$  & $2\pm{1}$  & NA\tabularnewline
 & NoP  & $144.82\pm{44.81}$  & $359\pm{253}$  & $72$  & $353$  & N/A & $100$  & N/A & $24$  & $60.3\pm{19.43}$  & $101$ \tabularnewline
\hline 
\multirow{6}{*}{\rotatebox{90}{Hyperplane}}  & CR  & $91.19\pm{3.28}$  & \textbf{$\textbf{92.26}\pm\textbf{2.67}$}  & $76\pm{8}$  & $86\pm{6}$  & $87\pm{4}$  & $74.78\pm{3.54}$  & $81.39\pm{2.2}$  & $91.65\pm{2.42}$  & $87.6\pm{6.2}$  & $90.45\pm{2}$ \tabularnewline
 & TrT  & $0.48\pm{0.007}$  & $0.15\pm{0.004}$  & N/A & N/A & N/A & $0.0004\pm{0.0001}$  & $1.91\pm{0.15}$  & $1.2\pm{0.2}$  & $0.4\pm{0.1}$  & $3.34\pm{1.97}$ \tabularnewline
 & TsT  & $0.002\pm{0.0003}$  & $0.02\pm{0.0013}$  & N/A & N/A & N/A & $0.0026\pm{0.0016}$  & $5.32\pm{0.44}$  & $0.6\pm{0.13}$  & $0.3\pm{0.03}$  & $1.67\pm{1}$ \tabularnewline
 & HN  & $16\pm{2.3}$  & $9.44\pm{1}$  & $12$  & $42$  & N/A & N/A & $100$  & $4.8\pm{2.4}$  & $2.76\pm{0.47}$  & $10$ \tabularnewline
 & HL  & $1$  & $1$  & $2$  & $3$  & $2$ & N/A & N/A & $2.4\pm{1.2}$  & $3\pm{2}$  & NA\tabularnewline
 & NoP  & $114\pm{18.8}$  & $69.1\pm{7}$  & $98$  & $0.5$K  & N/A & $120$  & N/A & $57.88\pm{28.7}$  & $54.68\pm{10.92}$ & $121$ \tabularnewline
\hline 
\multirow{6}{*}{\rotatebox{90}{Occupancy}} & CR  & $90.72\pm{15.9}$  & $87.97\pm{17.27}$  & $72\pm{34}$ & $71\pm{34}$ & \textbf{$\textbf{99}\pm\textbf{0.2}$}  & $56.65\pm{34}$  & $86.02\pm{15.2}$  & $89.30\pm{23.4}$  & $89.33\pm{24}$  & $94.65\pm{11}$ \tabularnewline
 & TrT  & $0.7\pm{0.4}$  & $0.29\pm{0.04}$  & N/A & N/A & N/A & $0.05\pm{0.01}$  & $1.5\pm{2.2}$  & $0.61\pm{0.28}$  & $0.34\pm{0.31}$  & $0.4\pm{0.2}$ \tabularnewline
 & TsT  & $0.004\pm{0.0004}$  & $0.03\pm{0.005}$  & N/A & N/A & N/A & $0.0069\pm{0.01}$  & $1\pm{0.5}$  & $0.44\pm{0.03}$  & $0.26\pm{0.1}$  & $0.15\pm{0.07}$ \tabularnewline
 & HN  & $35.68\pm{6.45}$  & $21.75\pm{11}$  & $20$  & $30$  & N/A & N/A & $100$  & $2.5\pm{1.2}$  & $2.3\pm{0.49}$  & $10$ \tabularnewline
 & HL  & $1$  & $1$  & $2$  & $3$  & $2$ & N/A & N/A & $2\pm{1.4}$  & $1.4\pm{0.5}$  & NA\tabularnewline
 & NoP  & $257.8\pm{117}$  & $177\pm{87}$  & $162$  & $302$  & N/A & $8$  & N/A & $30\pm{14}$  & $27.7\pm{0.48}$  & $8$ \tabularnewline
\hline 
\multirow{6}{*}{\rotatebox{90}{RFID Localization}} & CR  & $98.29\pm{6.4}$  & $98.66\pm{7}$  & $95\pm{10}$ & $66\pm{10}$ & $99.6\pm{0.2}$ & N/A & $\textbf{99.99}\pm\textbf{0.01}$  & $60.4\pm{6.7}$  & $60.9\pm{7.6}$  & $99.58\pm{0.98}$ \tabularnewline
 & TrT  & $0.5\pm{0.01}$  & $0.27\pm{0.03}$  & N/A & N/A & N/A & N/A & $0.46\pm{0.18}$  & $0.8\pm{0.14}$  & $1\pm{0.2}$  & $6.89\pm{3.71}$ \tabularnewline
 & TsT  & $0.003\pm{0.0008}$  & $0.05\pm{0.02}$  & N/A & N/A & N/A & N/A & $0.75\pm{0.55}$  & $0.3\pm{0.1}$  & $0.3\pm{0.04}$  & $3.44\pm{1.85}$ \tabularnewline
 & HN  & $63.64\pm{13.74}$  & $100\pm{10.82}$  & $2$  & $25$  & N/A & N/A & $100$  & $1.57\pm{0.65}$  & $1.31\pm{0.46}$  & $10$ \tabularnewline
 & HL  & $1$  & $1.6\pm{0.5}$  & $12$  & $3$  & $2$ & N/A & N/A & $2\pm{1}$  & $2\pm{0.8}$  & NA\tabularnewline
 & NoP  & $513\pm{111}$  & $(1\pm{0.7})$K  & $55$  & $232$  & N/A & N/A & N/A & $42.7\pm{22.48}$  & $43.73\pm{13.52}$  & $561$\tabularnewline
\hline 
\multirow{6}{*}{\rotatebox{90}{KDDCup 10\%}} & CR  & \textbf{$\textbf{99.84}\pm\textbf{0.16}$}  & $99.83\pm{0.2}$  & $99.6\pm{1}$  & $99\pm{1}$  & $97\pm{0.6}$ & $98.55\pm{0.53}$  & $99.5\pm{0.4}$  & $99.3\pm{0.4}$  & $96.7\pm{6}$  & N/A\tabularnewline
 & TrT  & $0.54\pm{0.02}$  & $0.09\pm{0.005}$  & N/A & N/A & N/A & $0.0024\pm{0.0005}$  & $0.62\pm{0.07}$  & $5\pm{0.3}$  & $0.6\pm{0.04}$  & N/A\tabularnewline
 & TsT  & $0.002\pm{0.001}$  & $0.002\pm{0.001}$  & N/A & N/A & N/A & $0.02\pm{0.006}$  & $0.97\pm{0.08}$  & $0.2\pm{0.01}$  & $0.25\pm{0.08}$  & N/A\tabularnewline
 & HN  & $34\pm{2}$  & $36\pm{2}$  & $60$  & $60$  & N/A & N/A & $100$  & $1$  & $1$  & N/A\tabularnewline
 & HL  & $1$  & $1$  & $2$  & $3$  & $2$ & N/A & N/A & $1$  & $1$  & N/A\tabularnewline
 & NoP  & $1.5\pm{0.1}$K  & $1.6K\pm{87}$  & $2$K  & $2$K  & N/A & $500$  & N/A & $12$  & $12$  & N/A\tabularnewline
\hline 
\multirow{6}{*}{\rotatebox{90}{HEPMASS 19\%}} & CR  & \textbf{$\textbf{83.39}\pm\textbf{2}$}  & $83.04\pm{1.8}$  & $76\pm{4}$  & $70\pm{4}$  & $78\pm{1}$  & $80.11\pm{98.21}$  & $78.3\pm{2.2}$  & ${82.6}\pm{1.9}$  & $82.3\pm{2.2}$  & N/A\tabularnewline
 & TrT  & $0.56\pm{0.04}$  & $0.13\pm{0.02}$  & N/A & N/A & N/A & ${0.002}\pm{0.0004}$  & $2.06\pm{0.2}$  & $6$  & $1.5\pm{0.2}$  & N/A\tabularnewline
 & TsT  & $0.004\pm{0.015}$  & $0.02\pm{0.01}$  & N/A & N/A & N/A & ${0.04}\pm{0.022}$  & $3.8\pm{0.3}$  & $7.5$  & $0.3\pm{0.03}$  & N/A\tabularnewline
 & HN  & $10.88\pm{0.5}$  & $99\pm{4.5}$  & $40$  & $18$  & N/A & N/A & $100$  & $2.01\pm{0.69}$  & $2.01\pm{0.69}$  & N/A\tabularnewline
 & HL  & $1$  & $2\pm{0.7}$  & $2$  & $3$  & $2$ & N/A & N/A & $2.01\pm{0.69}$  & $2.01\pm{0.69}$  & N/A\tabularnewline
 & NoP  & $340\pm{17}$  & $730\pm{398}$  & $1$K  & $324$  & N/A & $2$K  & N/A & $24.14\pm{8.23}$  & $24.14\pm{8.23}$  & N/A\tabularnewline
\bottomrule
\end{tabular}} 
\par\end{centering}
\end{table}
It is reported in Tables \ref{ranking} and \ref{result} that DEVDAN produces the highest classification rates in Rotated MNIST, MNIST, KDDCup and HEPMASS problems. It is observed that DEVDAN's numerical results are inferior to its counterparts in six problems: SEA, Hyperplane, RFID, Permuted MNIST, Forest Covertype and Occupancy. For the first three problems, however, the gap to the best performing method is statistically insignificant - around 1\% while outperforming the remainder of the consolidated algorithms. This finding is likely attributed to the noise-free nature of the two problems. Note that the use of noise injected mechanism is akin to regularization mechanism and thus incurs some loss albeit its evident benefits.

Separately, it is also observed that the execution time of DEVDAN is faster than other benchmarked algorithms except for ADL and Incremental Boosting in both training time and testing time, although it consists of a generative phase and discriminative phase.  In the realm of hidden node and network parameters, DEVDAN generates a comparable level of complexities compared to ADL in some cases. For instance, the structural learning mechanism contributes substantially to lower network parameters without compromising the predictive accuracy in the case of KDDCup and HEPMASS.

\begin{table*}
\caption{The classification rate of DEVDAN obtained using 5 consecutive runs. It is also presented the numerical result of DEVDAN-R.}
\centering{}\label{runs} \scalebox{0.7}{ %
\begin{tabular}{llrrrrr}
\toprule
Dataset  & Method  & \multicolumn{1}{c}{I} & \multicolumn{1}{c}{II} & \multicolumn{1}{c}{III} & \multicolumn{1}{c}{IV} & \multicolumn{1}{c}{V}\tabularnewline
\hline 
\multirow{2}{*}{Rotated MNIST}  & DEVDAN  & $76.48\pm9.66$ & $77.77\pm10.76$ & $77.37\pm10.82$ & $77.75\pm9.33$ & $78.09\pm9.32$\tabularnewline
 & DEVDAN-R  & NA & NA & NA & NA & NA\tabularnewline
\hline 
\multirow{2}{*}{Permuted MNIST}  & DEVDAN  & $78.72\pm13.41$ & $78.75\pm13.60$ & $77.45\pm13.46$ & $76.67\pm13.97$ & $78.60\pm13.77$\tabularnewline
 & DEVDAN-R  & NA & NA & NA & NA & NA\tabularnewline
\hline 
\multirow{2}{*}{MNIST}  & DEVDAN  & $87.49\pm6.25$ & $87.48\pm5.54$ & $87.11\pm5.77$ & $86.79\pm6.89$ & $86.12\pm7.83$\tabularnewline
 & DEVDAN-R  & NA & NA & NA & NA & NA\tabularnewline
\hline 
\multirow{2}{*}{Forest Covertype}  & DEVDAN  & $83.39\pm9.30$ & $83.54\pm9.29$ & $83.29\pm9.06$ & $83.54\pm9.59$ & $83.42\pm9.37$\tabularnewline
 & DEVDAN-R  & NA & NA & NA & NA & NA\tabularnewline
\hline 
\multirow{2}{*}{SEA}  & DEVDAN  & $91.12\pm7.12$ & $91.24\pm6.96$ & $92.10\pm6.59$ & $92.09\pm6.46$ & $91.60\pm6.12$\tabularnewline
 & DEVDAN-R  & $92.29\pm6.24$ & $92.38\pm6.21$ & $92.06\pm6.5$ & $92.46\pm6.02$ & $92.23\pm5.9$\tabularnewline
\hline 
\multirow{2}{*}{Hyperplane}  & DEVDAN  & $92.46\pm2.35$ & $91.44\pm3.1$ & $91.2\pm3.3$ & $92.4\pm2.87$ & $91.9\pm2.79$\tabularnewline
 & DEVDAN-R  & $92.53\pm2.61$ & $92.6\pm3.03$ & $92.67\pm1.9$ & $92.72\pm2.15$ & $92.69\pm1.85$\tabularnewline
\hline 
\multirow{2}{*}{Occupancy}  & DEVDAN  & $92.11\pm13.09$ & $90.72\pm15.92$ & $90.83\pm17.24$ & $90.72\pm17.15$ & $91.44\pm17.39$\tabularnewline
 & DEVDAN-R  & $72.26\pm31.13$ & $72.26\pm31.13$ & $72.26\pm31.13$ & $72.26\pm31.13$ & $72.26\pm31.13$\tabularnewline
\hline 
\multirow{2}{*}{RFID}  & DEVDAN  & $98.71\pm4.08$ & $98.52\pm4.65$ & $98.29\pm6.40$ & $98.52\pm5.06$ & $98.57\pm5.38$\tabularnewline
 & DEVDAN-R  & $96.91\pm7.74$ & $96.66\pm6.57$ & $95.42\pm10.7$ & $96.31\pm8.42$ & $93\pm15.49$\tabularnewline
\hline 
\multirow{2}{*}{KDDCup 10\%}  & DEVDAN  & $99.8390\pm0.1585$ & $99.8424\pm0.1483$ & $99.8498\pm0.1365$ & $99.8456\pm0.1647$ & $99.8488\pm0.1346$\tabularnewline
 & DEVDAN-R  & $99.8137\pm0.2054$ & $99.8319\pm0.1828$ & $99.8357\pm0.1770$ & $99.8351\pm0.1842$ & $99.8450\pm0.1483$\tabularnewline
\hline 
\multirow{2}{*}{HEPMASS 19\%}  & DEVDAN  & $83.39\pm2$ & $83.81\pm1.86$ & $83.90\pm2.12$ & $83.93\pm2.01$ & $83.79\pm1.89$\tabularnewline
 & DEVDAN-R  & $83.97\pm2.08$ & $83.8\pm1.97$ & $83.38\pm1.999$ & $83.61\pm2$ & $81.77\pm2$\tabularnewline
\bottomrule
\end{tabular}} 
\end{table*}
We next verify the effectiveness of DEVDAN compared to DEVDAN-R. For both methods, we follow the standard procedures outlined in Subsection \ref{implementationdetails}. Table \ref{runs} points out that our hidden node growing strategy where only $\mu_{Bias}^{min}$, $\sigma_{Bias}^{min}$ are reset to achieve better numerical results than when all parameters, $\mu_{Bias}^{t}$, $\sigma_{Bias}^{t}$, $\mu_{Bias}^{min}$ and $\sigma_{Bias}^{min}$ are reset (DEVDAN-R). Moreover, DEVDAN-R is unsuccessful while dealing with Rotated MNIST, Permuted MNIST, MNIST and Forest Covertype datasets as its hidden units keep growing uncontrollably.

\subsection{The Visualization of Learning Performance}
\begin{figure*}[!t]
	    \begin{centering}  \includegraphics[scale=0.45]{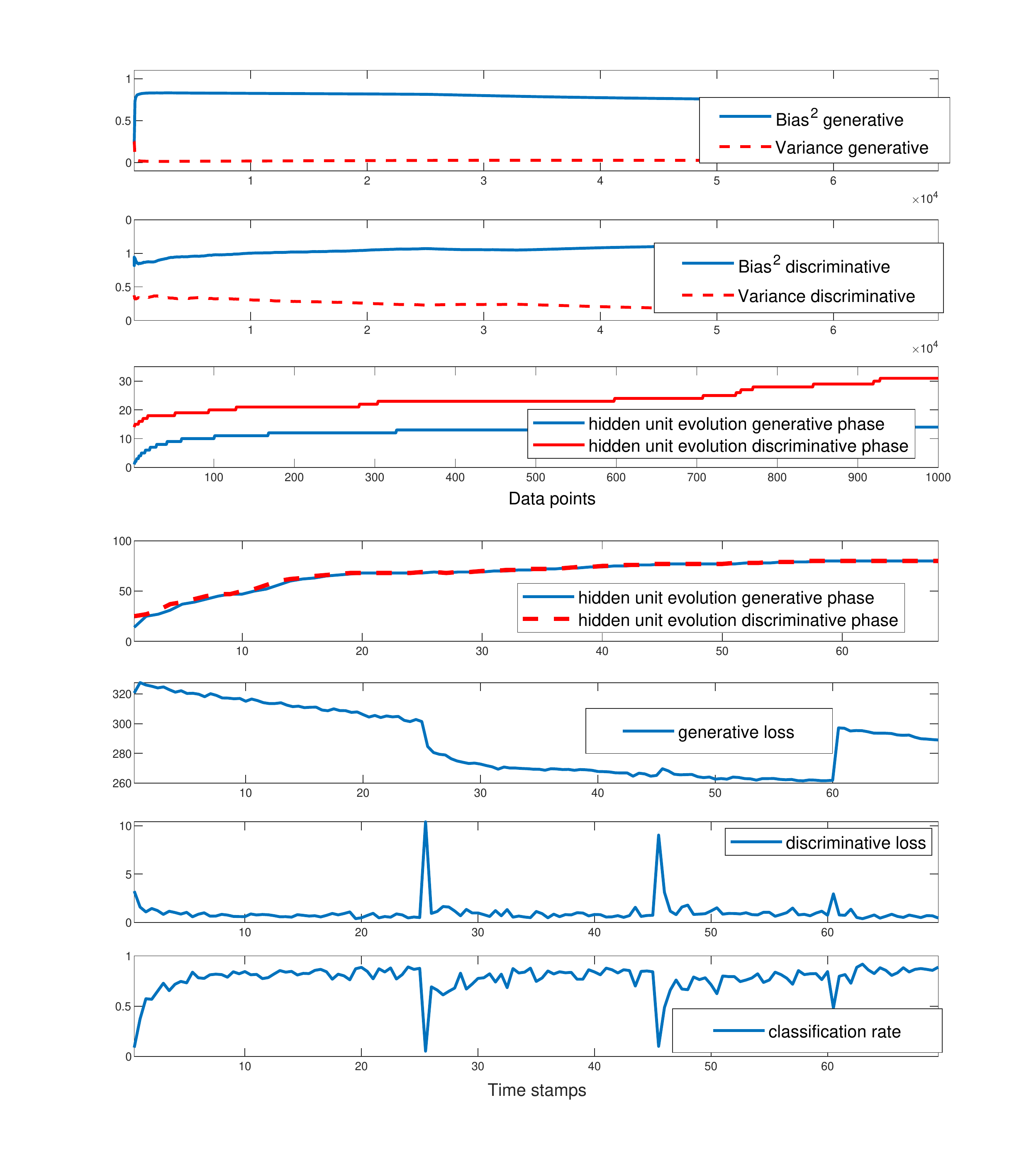}
	    \par\end{centering}
	    \caption{Performance metrics and hidden nodes evolution of Permuted MNIST problem.}
	    \label{fig:evolution}
\end{figure*}
\begin{figure*}[!t]
	    \begin{centering}  \includegraphics[scale=0.45]{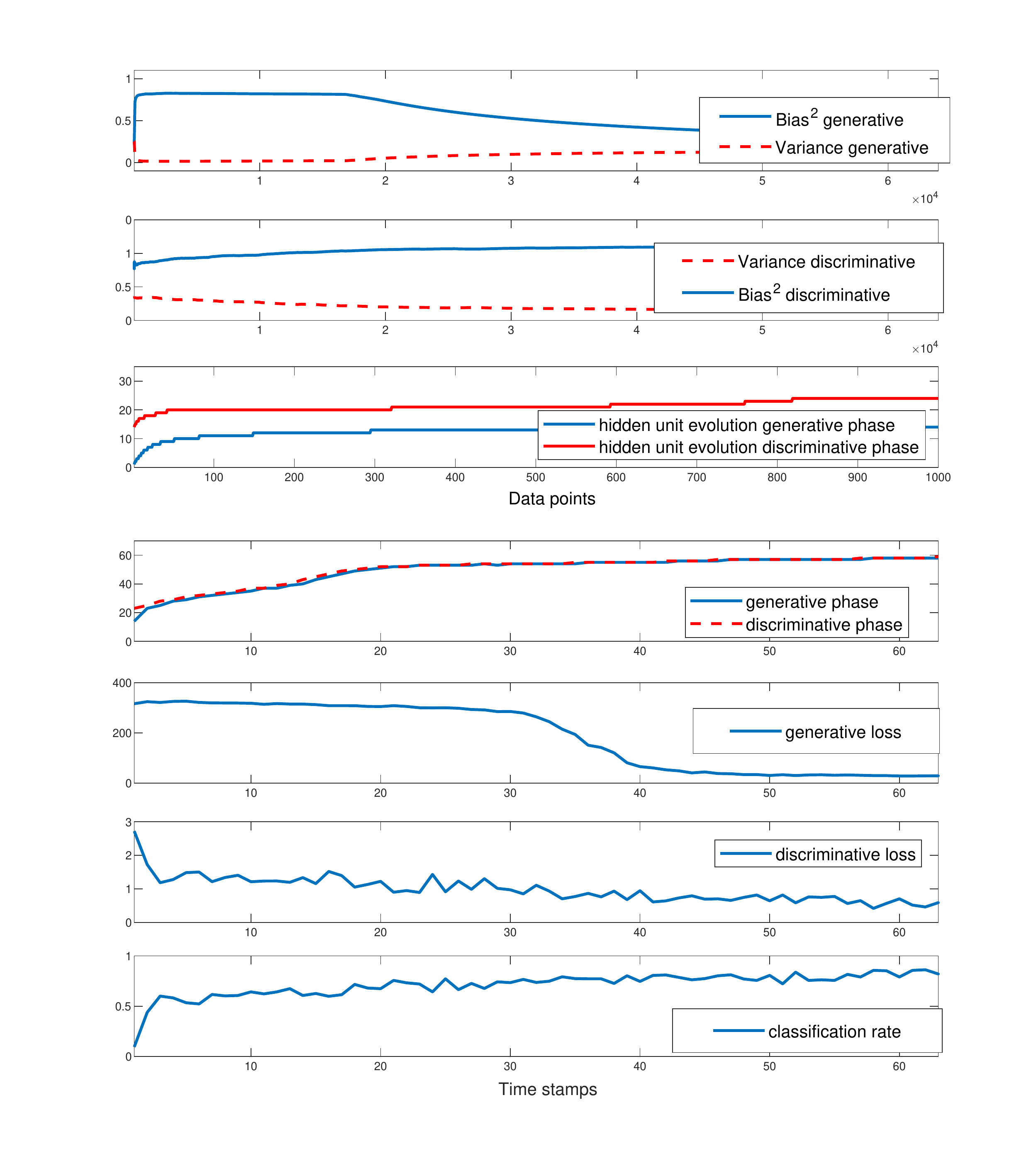}
	    \par\end{centering}
	    \caption{Performance metrics and hidden nodes evolution of Rotated MNIST problem.}
	    \label{fig:evolution2}
\end{figure*}

It is illustrated in Figs. \ref{fig:evolution} and \ref{fig:evolution2} that DEVDAN adopts a fully open and flexible structure where its structure is self-organized in both generative and discriminative phases. It is observed that a generative phase inherits a network structure constructed using unlabeled samples with respect to network reconstruction error aptitude. The discriminative phase further improves this network structure with access to the true class label. From the first three pictures in Figs. \ref{fig:evolution} and \ref{fig:evolution2}, the efficacy of the NS formula is demonstrated where hidden nodes can be timely added in the case of high bias and pruned in the case of high variance. This also empirically demonstrates the stability of the NS formula wherein each problem the NS formula is always able to find the appropriate network complexity for the given problem. Note that DEVDAN can be extended into a deep version with ease  by applying the greedy layer-wise learning process \cite{Bengio_Greedy} because the NS formula can be applied in every layer of a deep neural network.

From the last four pictures in Figs. \ref{fig:evolution} and \ref{fig:evolution2}, it is observed that the classification rate increases and the losses decrease as the number of nodes increases. It implies that the network capacity plays an important role to increase the predictive performance. Note that the sudden increase of loss in Fig. \ref{fig:evolution} (around $k=[25,45,60]$) indicates a strong presence of concept drift. This problem can be coped with the hidden unit growing and parameter adjustment mechanism where the discriminative loss rapidly decreases in the next time stamp. 
% It is also perceived that the addition of hidden units occurs when the network performance is compromised. The addition of hidden units leads to the increase of network variance and this issue is identified by the network pruning strategy by getting rid of the least contributing node.

\subsection{Statistical Test}
Numerical results of DEVDAN is statistically validated using the Wilcoxon signed-rank test \cite{woolson2007wilcoxon} to assess the numerical results of DEVDAN and other methods are significantly different. The Wilcoxon signed-rank test is used here because it supports a pairwise comparison of two different algorithms and is an alternative of the t-test for non normally distributed objects. The numerical evaluation is done by examining the residual error of predictive models. The rejection of the null hypothesis indicates that DEVDAN's predictive accuracy is significantly better than its counterpart. Incremental bagging, Incremental boosting, OMB and Learn++.NSE are excluded from our statistical test because its predictive output does not satisfy the partition of unity property leading to incomparable residual errors.
\begin{table*}[!t]
\caption{The wilcoxon signed-rank test result. The mark $\times$ indicates the rejection of the null hypothesis.}
\centering{}\label{wilcoxon} \scalebox{0.7}{ %
\begin{tabular}{lcccccccccc}
\toprule
 & \multicolumn{10}{c}{Dataset}\tabularnewline
\midrule 
 & R. MNIST  & P. MNIST & MNIST & F. Covertype & SEA  & Hyperplane & Occupancy & RFID & KDDCup $10\%$  & HEPMASS 19\%\tabularnewline
\midrule 
ADL  & $\times$  & $\times$  & $\times$  & $\times$  & $\times$  & $\times$  & $\times$  & $\times$  & $\times$  & $\times$ \tabularnewline
HAT & $\times$  & $\times$  & $\times$  & $\times$  & $\times$  & $\times$  & $\times$  & $\times$  & $\times$  & $\times$ \tabularnewline
PNN & $\times$  & $\times$  & $\times$  & $\times$  & $\times$  & $\times$  & $\times$  & $\times$  & $\times$  & $\times$ \tabularnewline
OMB & N/A & N/A & N/A & N/A & N/A & N/A & N/A & N/A & N/A & N/A\tabularnewline
$Inc_{Boosting}$  & N/A & N/A & N/A & N/A & N/A & N/A & N/A & N/A & N/A & N/A\tabularnewline
$Inc_{Bagging}$  & N/A & N/A & N/A & N/A & N/A & N/A & N/A & N/A & N/A & N/A\tabularnewline
pENsemble  & $\times$  & $\times$  & $\times$  & $\times$  & $\times$  & $\times$  & $\times$  & $\times$  & $\times$  & $\times$ \tabularnewline
pENsemble+ & $\times$  & $\times$  & $\times$  & $\times$  & $\times$  & $\times$  & $\times$  & $\times$  & $\times$  & $\times$ \tabularnewline
LEARN++.NSE & N/A & N/A & N/A & N/A & N/A & N/A & N/A & N/A & N/A & N/A\tabularnewline
\bottomrule
\end{tabular}} 
\end{table*}

Table \ref{ranking} sums up ranking of the accuracy of the ten algorithms, the effectiveness of DEVDAN is demonstrated where it outperforms the other nine algorithms in four of tens problems: Rotated MNIST, MNIST, KDDCup and HEPMASS. Table \ref{wilcoxon} summarizes the outcome of the statistical test. It is perceived from Table \ref{wilcoxon} that DEVDAN's predictive performance is significantly different from other algorithms and is statistically confirmed via the Wilcoxon signed-rank test that DEVDAN is better than other algorithms in those four problems: Rotated MNIST, MNIST, KDDCup and HEPMASS.

\subsection{Discussion}
Numerical results in Tables \ref{ranking}, where DEVDAN outperforms other methods, demonstrate that coupled-generative-discriminative training phases are capable of improving the predictive performance for data stream analytic with or without the label. This also exhibits that the evolution mechanism governed by NS formula and parameter learning strategies using the SGD method with momentum are stable while working together. On the other hand, the performance degradation in Forest Covertype and Permuted MNIST problems are suspected due to the real drift feature of the problem. This issue leads to the structural learning mechanism of the generative and discriminative phase to be not synchronized. That is, the virtual drift handling mechanism of the generative phase distracts the location of initial points for the discriminative phase.

In terms of time complexity, DEVDAN computation time is comparable to ADL and even faster than other methods in all cases. This confirms that DEVDAN's time complexity is linear and fairly low as estimated in (\ref{timecom}). It is faster than HAT and PNN because it arrives at a less complex network structure than them. Several methods are crafted from the concept of ensemble models which comprises multiple classifiers, thereby being computationally more expensive than DEVDAN where the adaptive and evolving characteristic is realized in the hidden node level. Although pENsemble and pENsemble+ evolve a lower number of base classifiers than DEVDAN, it incurs slower training and testing times than DEVDAN because it adopts the ensemble concept. 

From Tables \ref{ranking} and \ref{runs}, it can be noticed that DEVDAN is consistent while delivering predictive performance. Moreover, other results in Table \ref{result} show that DEVDAN outperforms ADL, HAT, and PNN in Rotated MNIST and MNIST, Forest Covertype, Occupancy, KDDCup and HEPMASS datasets, although DEVDAN is a single hidden layer network. These results exhibit the benefit of generative training phase as an unsupervised pretraining mechanism. Note that the parameter initialization of DNN possibly has a significant regularizing effect on the predictive model. Moreover, DNN training is non-deterministic and ends up to a different function in every execution. Having an unsupervised pretraining mechanism enables a DNN to consistently halt in the same region of function space. The region, where an unsupervised pretraining mechanism is utilized, is smaller implying that this mechanism decreases the estimation process variance, which decreases the risk of overfitting. In other words, generative training phase initializes DEVDAN's parameters into an inescapable region \cite{goodfellow2016deep} and consequently, the performance is more consistent and more likely to be good than without this phase. This result also confirms our hypothesis that the generative training phase is capable of improving predictive performance for data stream analytic exploiting unlabeled data.
%These results are understood because DEVDAN utilizes generative training phase as an unsupervised pretraining mechanism.
%with the absence of the true class label.
%Having an unsupervised pretraining mechanism can initialize DEVDAN's parameters into the inaccessible region. This mechanism enables a DNN to consistently halt in the same region of function space \cite{goodfellow2016deep}.
%can initialize DEVDAN's parameters into the same region of function space.
%In other words, generative training initializes DEVDAN 

The comparison between DEVDAN and DEVDAN-R outlined in Table \ref{runs} suggests that resetting $\mu_{Bias}^{min}$, $\sigma_{Bias}^{min}$ and preserving $\mu_{Bias}^{t}$, $\sigma_{Bias}^{t}$ are the key to control the stability of the network evolution. This facet can be understood from the spirit of the NS formula derived from integral approximation over all input space to better reflect the true data distribution. Setting $\mu_{Bias}^{t}$, $\sigma_{Bias}^{t}$ to zero during the addition of a new hidden unit causes loss of information of preceding samples. This fact is confirmed from the concept of the hidden unit in a neural network which differs from the concept of the hidden unit in the RBF network where every unit represents a particular input space partition. The hidden unit contribution in DNN is judged from its aptitude to drive the error to zero. Furthermore, by simply resetting $\mu_{Bias}^{min}$, $ \sigma_{Bias}^{min}$ it is capable of finding a new level after previous drift - a very important aspect of concept drift detection.

\subsection{Ablation Study}
Since DEVDAN consists of several learning mechanisms, it has a good deal in common with existing methods in the literature. As a result, we conduct ablation study by removing or adding components in order to provide additional insight into the effect of each DEVDAN's learning mechanism. Specifically, we measure the effect of generative training phase, hidden unit growing and pruning mechanisms.

\begin{table}[!t]
\caption{Ablation study results. Scenario: A) without generative training phase, B) without hidden
node growing mechanism, C) without hidden node pruning mechanism.}
\begin{centering}
\label{result-1-1-1-1-1} \scalebox{0.8}{ %
\begin{tabular}{llrrrr}
\toprule 
 &  & DEVDAN & Scenario A & Scenario B & Scenario C\tabularnewline
\midrule 
\multirow{6}{*}{\rotatebox{90}{Rotated MNIST}} & CR  & $\textbf{76.48}\pm{\textbf{9.7}}$  & $76.19\pm{9.23}$  & $71.81\pm{5.38}$  & $74.87\pm{10.63}$ \tabularnewline
 & TrT  & $1.6\pm{0.13}$  & $0.43\pm{0.03}$  & $1.72\pm{0.07}$  & $2.12\pm{0.09}$ \tabularnewline
 & TsT  & $0.006\pm{0.001}$  & $0.008\pm{0.002}$  & $0.007\pm{0.0005}$  & $0.008\pm{0.001}$ \tabularnewline
 & HN  & $48.7\pm{9}$  & $26.47\pm{10}$  & $10$  & $52.13\pm{10.82}$ \tabularnewline
 & HL  & $1$  & $1$  & $1$  & $1$ \tabularnewline
 & NoP  & $(38\pm{8})$K  & $(20\pm{8.3})$K  & $8$K  & $(41\pm{9.8})$K \tabularnewline
\midrule 
\multirow{6}{*}{\rotatebox{90}{Forest Covertype}} & CR  & $\textbf{83.29}\pm \textbf{9.06}$  & $82.79\pm{9.5}$  & $79.06\pm10.93$  & $83.12\pm{9.42}$ \tabularnewline
 & TrT  & $0.6\pm{0.04}$  & $0.2\pm{0.02}$  & $0.56\pm{0.02}$  & $0.63\pm{0.04}$ \tabularnewline
 & TsT  & $0.005\pm{0.002}$  & $0.005\pm{0.002}$  & $0.0045\pm{0.002}$  & $0.005\pm{0.002}$ \tabularnewline
 & HN  & $70.78\pm{19.5}$  & $57.71\pm{17.89}$  & $7$  & $70.61\pm{20}$ \tabularnewline
 & HL  & $1$  & $1$  & $1$  & $1$ \tabularnewline
 & NoP  & $(4.4\pm{1.2})$K  & $(3.53\pm{1.11})$K  & $442$  & $(4.4\pm{1.2})$K \tabularnewline
\bottomrule
\end{tabular}} 
\par\end{centering}
\end{table}

The ablation study is carried out on Rotated MNIST and Forest Covertype datasets; the results are presented in Table \ref{result-1-1-1-1-1}. Generally, it is found that each component contributes to DEVDAN’s performance, with the most dramatic difference in the without-growing-mechanism scenario. This is understood because the hidden unit growing mechanism enables DEVDAN to increase its network capacity in respect to data distribution. It is obvious that having more network capacity helps to improve the predictive performance especially when the function to be learned is extremely complicated \cite{goodfellow2016deep}.

From Table \ref{result-1-1-1-1-1}, it is observed that the generative training phase contributes around 1\% improvement in terms of classification rate. This signifies that the generative phase helps to refine the predictive performance using unlabeled data as it can initialize the network parameters into the region that they do not escape \cite{goodfellow2016deep}. Meanwhile, disabling the hidden unit pruning mechanism increases the network complexity. This is evidenced by the number of created hidden units. As a result, it raises the risk of being suffered from high variance dilemma. This is confirmed by the classification rates of the without-pruning-mechanism scenario where those have higher standard deviation compared to the results in Table \ref{result}. Moreover, DEVDAN's performances decrease by up to 2\% without hidden unit pruning mechanism.

\subsection{Limited Access to The Ground Truth} \label{latgt}
In the real-world environment, we may have limited access to the ground truth which causes the number of labeled data is less than the number of unlabeled data. An experiment simulating this scenario, also known as semi-supervised learning, is conducted to measure our approach's ability to generalize. This scenario is carried out on Rotated MNIST and Forest Covertype problems. In this experiment, we vary the portion of labeled data from 25\%, 50\%, to 75\% of the total data in a batch $T$. Two strategies are conducted to select the labeled data from a data batch. The first strategy is by random selection, whereas the second strategy is to effectively select the data using sample selection mechanism. 
\begin{table}[!t]
\caption{The numerical result of the limited-access-to-the-ground-truth scenario. The labeled data are selected randomly from each data batch.}
\centering{}\label{result-1-1-3} \scalebox{0.7}{ %
\begin{tabular}{llrrrrrrrr}
\toprule 
 &  & \multicolumn{2}{c}{$25\%$ labeled data } & \multicolumn{2}{c}{$50\%$ labeled data} & \multicolumn{2}{c}{$75\%$ labeled data} & \multicolumn{2}{c}{$100\%$ labeled data}\tabularnewline
\midrule 
 &  & DEVDAN  & ADL  & DEVDAN  & ADL  & DEVDAN  & ADL  & DEVDAN  & ADL\tabularnewline
\midrule 
\multirow{6}{*}{\rotatebox{90}{Rotated MNIST}} & CR  & \textbf{$\textbf{61.81}\pm\textbf{12.87}$}  & $61.25\pm{10.9}$  & \textbf{$\textbf{69.66}\pm\textbf{10.53}$}  & $54.25\pm{7.6}$  & \textbf{$\textbf{72.59}\pm\textbf{10.42}$}  & $69.62\pm{0.08}$  & \textbf{$\textbf{76.48}\pm\textbf{9.7}$}  & $73.97\pm{9.92}$ \tabularnewline
 & TrT  & $1.2\pm{0.03}$  & $0.165\pm{0.46}$  & $1.3\pm{0.04}$  & $0.2\pm{0.06}$  & $1.45\pm{0.08}$  & $0.33\pm{0.09}$  & $1.6\pm{0.13}$  & $0.38\pm{0.41}$ \tabularnewline
 & TsT  & $0.006\pm{0.001}$  & $0.059\pm{0.02}$  & $0.006\pm{0.001}$  & $0.09\pm{0.04}$  & $0.006\pm{0.001}$  & $0.07\pm{0.09}$  & $0.006\pm{0.001}$  & $0.06\pm{0.02}$ \tabularnewline
 & HN  & $34\pm{5.3}$  & $45.53\pm{1.7}$  & $40\pm{4.5}$  & $40\pm{0.45}$  & $59\pm{13}$  & $50.32\pm{2.9}$  & $48.7\pm{9}$  & $66.98\pm{9.3}$ \tabularnewline
 & HL  & $1$  & $1.1\pm{0.31}$  & $1$  & $3.3\pm{1.56}$  & $1$  & $1.14\pm{0.35}$  & $1$  & $1.14\pm{0.4}$ \tabularnewline
 & NoP  & $(25\pm{6.4})$K  & $(10\pm{1.4})$K  & $(30\pm{5.6})$K  & $(6.5\pm{0.9})$K  & $(43\pm{11})$K  & $(9\pm{2.4})$K  & $(38\pm{8})$K  & $(18\pm{7.5})$K \tabularnewline
\midrule 
\multirow{6}{*}{\rotatebox{90}{Forest Covertype}} & CR  & \textbf{$\textbf{78.69}\pm\textbf{10.59}$}  & $57.77\pm{18.35}$  & \textbf{$\textbf{80.88}\pm\textbf{9.9}$}  & $66.65\pm{14.49}$  & \textbf{$\textbf{82.21}\pm\textbf{9.27}$}  & $66.43\pm{13.09}$  & \textbf{$\textbf{83.29}\pm\textbf{9.06}$}  & $81.97\pm22.47$ \tabularnewline
 & TrT  & $0.43\pm{0.02}$  & $1.6\pm{1.5}$  & $0.48\pm{0.03}$  & $1.2\pm{1.1}$  & $0.54\pm{0.04}$  & $1.5\pm{1.4}$  & $0.6\pm{0.04}$  & $0.17\pm{0.01}$ \tabularnewline
 & TsT  & $0.004\pm{0.001}$  & $0.9\pm{0.9}$  & $0.004\pm{0.002}$  & $0.8\pm{0.5}$  & $0.004\pm{0.002}$  & $0.8\pm{0.5}$  & $0.005\pm{0.002}$  & $0.02\pm{0.003}$ \tabularnewline
 & HN  & $65.15\pm{12.9}$  & $939\pm{13.5}$  & $73.8\pm{13.57}$  & $836\pm{19}$  & $69.49\pm{12.53}$  & $902\pm{45}$  & $70.78\pm{19.5}$  & $20\pm{10}$ \tabularnewline
 & HL  & $1$  & $53.52\pm{39}$  & $1$  & $46.83\pm{29.5}$  & $1$  & $50.35\pm{30}$  & $1$  & $1$ \tabularnewline
 & NoP  & $(3.7\pm{0.9})$K  & $(8.7\pm{4.7})$K  & $(4.1\pm{1})$K  & $(10\pm{4.8})$K  & $(4.1\pm{0.8})$K  & $(13\pm{6})$K  & $(4.4\pm{1.2})$K  & $159\pm{81}$ \tabularnewline
\bottomrule
\end{tabular}} 
\end{table}
%All values are obtained from 5 consecutive runs.

The numerical results of the first strategy are tabulated in Table \ref{result-1-1-3}. We compare our method against the second best performant, ADL. It can be observed that DEVDAN obtains the best classification rate, significantly outperforming ADL. For instances, the difference is about 0.5\% to 20\% in terms of the classification rate. Interestingly, DEVDAN's performance on Forest Covertype problem having 75\% labeled data is better than ADL for every labeled data amount considered (Table \ref{result}). This result is understood as the unlabeled data are exploited by generative training phase. The robust features extracted by generative training phase may help the discriminative training phase to perform better. As a result, we may expect the generative phase to improve the performance when the number of unlabeled data is greater than the number of labeled data \cite{goodfellow2016deep}.

In the second strategy, a sample selection mechanism is employed to select the data which is useful for the discriminative training phase. To emphasize the importance of this, consider the following scenario: In the real-world situation, the experts may not be able to label all the incoming data. It is required to label several data which may help to improve the classification performance. Intuitively, this data should be a difficult sample. That is, a data sample which is geometrically close to the decision boundary separating between classes. The following formula from \cite{pensembleplus} is implemented in this experiment as a sample selection mechanism as per in (\ref{conflevel}): 
\begin{equation}
    conf=\frac{y_1}{y_1+y_2}<\delta\label{conflevel}
\end{equation}
where $y_1$ and $y_2$ are the highest and the second highest multiclass probability, and $\delta$ is the minimum confidence level. In other words, the ground truth is only revealed to the sample whose $conf$ is less than $\delta$. In this experiment, the value of $\delta$ is selected: 0.7.
\begin{table}[!t]
\caption{The numerical result of the limited-access-to-the-ground-truth scenario. The labeled data are selected using a sample selection mechanism from each data batch.}
\begin{centering}
\label{result-1-1-2} \scalebox{0.8}{ %
\begin{tabular}{llrrrr}
\toprule
 &  & $25\%
$ labeled data  & $50\%
$ labeled data & $75\%
$ labeled data & $100\%$ labeled data\tabularnewline
\hline 
\multirow{6}{*}{\rotatebox{90}{Rotated MNIST}} & CR  & $62.96\pm{12.88}$  & $69.87\pm{13.18}$  & $74.12\pm{10.58}$  & \textbf{$76.48\pm9.7$} \tabularnewline
 & TrT  & $1.24\pm{0.07}$  & $1.33\pm{0.04}$  & $1.4\pm{0.05}$  & $1.6\pm{0.13}$ \tabularnewline
 & TsT  & $0.007\pm{0.002}$  & $0.007\pm{0.001}$  & $0.006\pm{0.001}$  & $0.006\pm{0.001}$ \tabularnewline
 & HN  & $45.45\pm{6.2}$  & $45.12\pm{6.6}$  & $44.05\pm{8.7}$  & $48.7\pm{9}$ \tabularnewline
 & HL  & $1$  & $1$  & $1$  & $1$ \tabularnewline
 & NoP  & $(33\pm{8})$K  & $(33\pm{7})$K  & $(33\pm{9})$K  & $(38\pm{8})$K \tabularnewline
\hline 
\multirow{6}{*}{\rotatebox{90}{Forest Covertype}} & CR  & $73.96\pm{11.95}$  & $76.66\pm{10.93}$  & $78.05\pm{10.92}$  & \textbf{$83.29\pm9.06$} \tabularnewline
 & TrT  & $0.42\pm{0.02}$  & $0.47\pm{0.02}$  & $0.52\pm{0.04}$  & $0.6\pm{0.04}$ \tabularnewline
 & TsT  & $0.004\pm{0.001}$  & $0.004\pm{0.001}$  & $0.004\pm{0.001}$  & $0.005\pm{0.002}$ \tabularnewline
 & HN  & $54.19\pm{10.05}$  & $57.39\pm{9.4}$  & $59.54\pm{15.67}$  & $70.78\pm{19.5}$ \tabularnewline
 & HL  & $1$  & $1$  & $1$  & $1$ \tabularnewline
 & NoP  & $(3.1\pm{0.7})$K  & $(3.2\pm{0.7})$K  & $(3.4\pm{1})$K  & $(4.4\pm{1.2})$K \tabularnewline
\bottomrule
\end{tabular}} 
\par\end{centering}
\end{table}

The tabulated numerical results in Table \ref{result-1-1-2} show that the sample selection mechanism helps DEVDAN to achieve around 0.2\% to 2\% improvement in terms of accuracy in Rotated MNIST problem. This is reasonable as exploiting the labeled data which is situated around the decision boundary increases the network's confidence while executing the classification task. On the other hand, the second strategy decreases DEVDAN's classification rate in Forest Covertype problem. Note that this dataset has imbalance class proportion where five out of seven classes have less than 7\% proportion. One should be very careful while implementing sample selection in this situation, the majority class samples of a particular batch are not always included in the discriminative training as its $conf$ has already been greater than $\delta$. As a result, DEVDAN is unable to refine its predictive performance using those samples which decreases the accuracy in the next data batch. All in all, these experiments suggest that DEVDAN is potentially able to handle semi-supervised learning problem. This is reasonable as its coupled-generative-discriminative training phases are capable of handling the concept drift with or without the label.

%From Table \ref{ranking}, the effectiveness of DEVDAN is demonstrated where it outperforms other nine algorithms in four of tens problems: Rotated MNIST, MNIST, KDDCup and HEPMASS. The statistical test is undertaken in all problems to confirm whether DEVDAN produces significantly better accuracy than other algorithms in all problems. No comparison is made against the remainder of banchmarked algorithms, Learn++, Learn++NSE, Incremental Bagging and Incremental Boosting, because their predictive outputs do not range in $[0,1]$ leading to incomparable residual errors. It is perceived from Table \ref{wilcoxon} that DEVDAN's predictive performance is significantly different from those of pENsemble, pENsemble+, AE and DAE statistically confirmed via the Wilcoxon signed-rank test.

%Numerical results of DEVDAN is statistically validated using the Wilcoxon signed-rank test \cite{woolson2007wilcoxon} to assess whether difference in numerical results of DEVDAN and other methods are significantly different. The Wilcoxon signed-rank test is used here because it supports a pairwise comparison of two different algorithms and is an alternative of the t-test for non normally distributed objects. The numerical evaluation is done by examining the residual error of predictive models. The rejection of the null hypothesis indicates that DEVDAN's predictive accuracy is significantly better than its counterpart. Table \ref{ranking} sums up ranking of accuracy of the ten algorithms while Table \ref{wilcoxon} summarizes the outcome of statistical test.

\section{Conclusion}
We introduced DEVDAN, an evolving denoising autoencoder which combines generative and discriminative training phases for data stream analytic. DEVDAN features an open structure both in the generative phase and in the discriminative phase where input features can be automatically added and discarded on the fly. Furthermore, DEVDAN is free of the problem-specific threshold and works fully in the single-pass learning fashion. Through extensive experiments, we found that DEVDAN significantly outperforms other algorithms in 4 of 10 problems in the supervised learning scenario and in all problems in the semi-supervised learning scenario. This fact also supports the relevance of the generative phase for online data stream which contributes toward the refinement of network structure in an unsupervised fashion. In future work, we are interested in investigating the deep version of DEVDAN to increase its generalization power. A further point, the ideas from state-of-the-arts semi-supervised learning literature will be incorporated to find the effective algorithm for semi-supervised learning. Separately, we are also interested in exploring the effectiveness of DEVDAN for transfer learning application. To allow one to reproduce our numerical results, the source codes of DEVDAN can be accessed from \url{https://bit.ly/2Jk3Pzf}. A short video also provided in the link demonstrating DEVDAN's learning mechanism.

\bibliographystyle{unsrt}  

 \bibliography{references}

\begin{thebibliography}{10}
\expandafter\ifx\csname url\endcsname\relax
  \def\url#1{\texttt{#1}}\fi
\expandafter\ifx\csname urlprefix\endcsname\relax\def\urlprefix{URL }\fi
\expandafter\ifx\csname href\endcsname\relax
  \def\href#1#2{#2} \def\path#1{#1}\fi

\bibitem{DeepExpandable}
J.~Yoon, E.~Yang, J.~Lee, S.~J. Hwang, Lifelong learning with dynamically
  expandable networks (2018).

\bibitem{IMM2012}
R.~B. Palm, Prediction as a candidate for learning deep hierarchical models of
  data, Master's thesis (2012).

\bibitem{LearningTheNumber}
J.~M. Alvares, M.~Salzmann, Learning the number of neurons in deep networks,
  in: D.~D. Lee, M.~Sugiyama, U.~V. Luxburg, I.~Guyon, R.~Garnett (Eds.),
  Advances in Neural Information Processing Systems 29, Curran Associates,
  Inc., 2016, pp. 2270--2278 (2016).

\bibitem{parameterprediction}
M.~Denil, B.~Shakibi, L.~Dinh, M.~Ranzato, N.~de~Freitas, Predicting parameters
  in deep learning, in: Proceedings of the 26th International Conference on
  Neural Information Processing Systems - Volume 2, NIPS'13, Curran Associates
  Inc., USA, 2013, pp. 2148--2156 (2013).

\bibitem{distilling}
G.~{Hinton}, O.~{Vinyals}, J.~{Dean}, {Distilling the Knowledge in a Neural
  Network}, ArXiv e-prints (Mar. 2015).
\newblock \href {http://arxiv.org/abs/1503.02531} {\path{arXiv:1503.02531}}.

\bibitem{GamaDataStream}
J.~Gama, Knowledge Discovery from Data Streams, 1st Edition, Chapman \&
  Hall/CRC, 2010 (2010).

\bibitem{Bengio_2013}
Y.~Bengio, A.~Courville, P.~Vincent, Representation learning: A review and new
  perspectives, IEEE Trans. Pattern Anal. Mach. Intell. 35~(8) (2013)
  1798--1828 (Aug. 2013).

\bibitem{learning_dynamic_AE}
A.~Pretorius, S.~Kroon, H.~Kamper,
  \href{http://proceedings.mlr.press/v80/pretorius18a.html}{Learning dynamics
  of linear denoising autoencoders}, in: J.~Dy, A.~Krause (Eds.), Proceedings
  of the 35th International Conference on Machine Learning, Vol.~80 of
  Proceedings of Machine Learning Research, PMLR, Stockholmsmässan, Stockholm
  Sweden, 2018, pp. 4141--4150 (10--15 Jul 2018).
\newline\urlprefix\url{http://proceedings.mlr.press/v80/pretorius18a.html}

\bibitem{zeng2018facial}
N.~Zeng, H.~Zhang, B.~Song, W.~Liu, Y.~Li, A.~M. Dobaie, Facial expression
  recognition via learning deep sparse autoencoders, Neurocomputing 273 (2018)
  643--649 (2018).

\bibitem{DEEPIOT}
M.~Mohammadi, A.~I. Al-Fuqaha, S.~Sorour, M.~Guizani, Deep learning for iot big
  data and streaming analytics: A survey, CoRR abs/1712.04301 (2017).

\bibitem{Zhou_incrementallearning}
G.~Zhou, K.~Sohn, H.~Lee, Online incremental feature learning with denoising
  autoencoders, Journal of Machine Learning Research 22 (2012) 1453--1461
  (2012).

\bibitem{VincentDAE}
P.~Vincent, H.~Larochelle, Y.~Bengio, P.-A. Manzagol, Extracting and composing
  robust features with denoising autoencoders, in: Proceedings of the 25th
  International Conference on Machine Learning, ICML '08, ACM, New York, NY,
  USA, 2008, pp. 1096--1103 (2008).

\bibitem{OnlineDeepLearning}
D.~Sahoo, Q.~D. Pham, J.~Lu, S.~C. Hoi, Online deep learning: Learning deep
  neural networks on the fly, arXiv preprint arXiv:1711.03705 abs/1711.03705
  (2017).

\bibitem{progressive}
A.~A. Rusu, N.~C. Rabinowitz, G.~Desjardins, H.~Soyer, J.~Kirkpatrick,
  K.~Kavukcuoglu, R.~Pascanu, R.~Hadsell,
  \href{http://arxiv.org/abs/1606.04671}{Progressive neural networks}, CoRR
  abs/1606.04671 (2016).
\newblock \href {http://arxiv.org/abs/1606.04671} {\path{arXiv:1606.04671}}.
\newline\urlprefix\url{http://arxiv.org/abs/1606.04671}

\bibitem{ADL}
A.~Ashfahani, M.~Pratama, Autonomous Deep Learning: Continual Learning Approach
  for Dynamic Environments, Society for Industrial and Applied Mathematics,
  2019, pp. 666--674 (2019).

\bibitem{rotatedMNIST}
D.~Lopez-Paz, M.~A. Ranzato, \href{https://bit.ly/2HkaaZr}{Gradient episodic
  memory for continual learning}, in: I.~Guyon, U.~V. Luxburg, S.~Bengio,
  H.~Wallach, R.~Fergus, S.~Vishwanathan, R.~Garnett (Eds.), Advances in Neural
  Information Processing Systems 30, Curran Associates, Inc., 2017, pp.
  6467--6476 (2017).
\newline\urlprefix\url{https://bit.ly/2HkaaZr}

\bibitem{permuttedMNIST}
R.~K. Srivastava, J.~Masci, S.~Kazerounian, F.~Gomez, J.~Schmidhuber,
  \href{http://papers.nips.cc/paper/5059-compete-to-compute.pdf}{Compete to
  compute}, in: C.~J.~C. Burges, L.~Bottou, M.~Welling, Z.~Ghahramani, K.~Q.
  Weinberger (Eds.), Advances in Neural Information Processing Systems 26,
  Curran Associates, Inc., 2013, pp. 2310--2318 (2013).
\newline\urlprefix\url{http://papers.nips.cc/paper/5059-compete-to-compute.pdf}

\bibitem{lecun-mnisthandwrittendigit-2010}
Y.~LeCun, C.~Cortes, \href{http://yann.lecun.com/exdb/mnist/}{{MNIST}
  handwritten digit database} (2010) [cited 2016-01-14 14:24:11].
\newline\urlprefix\url{http://yann.lecun.com/exdb/mnist/}

\bibitem{blackard1999comparative}
J.~A. Blackard, D.~J. Dean, Comparative accuracies of artificial neural
  networks and discriminant analysis in predicting forest cover types from
  cartographic variables, Computers and Electronics in Agriculture vol.24
  (1999) 131--151 (1999).

\bibitem{SEA}
W.~N. Street, Y.-S. Kim, \href{http://doi.acm.org/10.1145/502512.502568}{A
  streaming ensemble algorithm (sea) for large-scale classification}, in:
  Proceedings of the Seventh ACM SIGKDD International Conference on Knowledge
  Discovery and Data Mining, KDD '01, ACM, New York, NY, USA, 2001, pp.
  377--382 (2001).
\newblock \href {https://doi.org/10.1145/502512.502568}
  {\path{doi:10.1145/502512.502568}}.
\newline\urlprefix\url{http://doi.acm.org/10.1145/502512.502568}

\bibitem{MOA}
A.~Bifet, G.~Holmes, R.~Kirkby, B.~Pfahringer,
  \href{http://dl.acm.org/citation.cfm?id=1756006.1859903}{Moa: Massive online
  analysis}, J. Mach. Learn. Res. 11 (2010) 1601--1604 (Aug. 2010).
\newline\urlprefix\url{http://dl.acm.org/citation.cfm?id=1756006.1859903}

\bibitem{candanedo2016accurate}
L.~M. Candanedo, V.~Feldheim, Accurate occupancy detection of an office room
  from light, temperature, humidity and co2 measurements using statistical
  learning models, Energy and Buildings 112 (2016) 28--39 (2016).

\bibitem{rfidL}
A.~Ashfahani, M.~Pratama, E.~Lughofer, Q.~Cai, H.~Sheng, An Online RFID
  Localization in the Manufacturing Shopfloor, Springer International
  Publishing, 2019, pp. 287--309 (2019).
\newblock \href {https://doi.org/10.1007/978-3-030-05645-2_10}
  {\path{doi:10.1007/978-3-030-05645-2_10}}.

\bibitem{KDDCup}
S.~J. Stolfo, W.~Fan, W.~Lee, A.~Prodromidis, P.~K. Chan, Cost-based modeling
  for fraud and intrusion detection: Results from the jam project, in: In
  Proceedings of the 2000 DARPA Information Survivability Conference and
  Exposition, IEEE Computer Press, 2000, pp. 130--144 (2000).

\bibitem{Baldi2014SearchingFE}
P.~Baldi, P.~D. Sadowski, D.~Whiteson, Searching for exotic particles in
  high-energy physics with deep learning., Nature communications 5 (2014) 4308
  (2014).

\bibitem{pENsemble}
M.~Pratama, W.~Pedrycz, E.~Lughofer, Evolving ensemble fuzzy classifier, IEEE
  Transactions on Fuzzy Systems (2018) 1--1 (2018).

\bibitem{jung2017online}
Y.~H. Jung, J.~Goetz, A.~Tewari, Online multiclass boosting, in: Advances in
  neural information processing systems, 2017, pp. 919--928 (2017).

\bibitem{IncBaggingBoosting}
C.~Oza~Nikunj, J.~Russell~Stuart, Online bagging and boosting. jaakkola tommi
  and richardson thomas, editors, in: Eighth International Workshop on
  Artificial Intelligence and Statistics, 2001, pp. 105--112 (2001).

\bibitem{pensembleplus}
M.~Pratama, E.~Dimla, E.~Lughofer, W.~Pedrycz, T.~Tjahjowidodo,
  \href{http://arxiv.org/abs/1711.01843}{Online tool condition monitoring based
  on parsimonious ensemble+}, CoRR abs/1711.01843 (2017).
\newblock \href {http://arxiv.org/abs/1711.01843} {\path{arXiv:1711.01843}}.
\newline\urlprefix\url{http://arxiv.org/abs/1711.01843}

\bibitem{Learn++NSE}
R.~Elwell, R.~Polikar, Incremental learning of concept drift in nonstationary
  environments, Trans. Neur. Netw. 22~(10) (2011) 1517--1531 (Oct. 2011).

\bibitem{kirkpatrick2016overcoming}
J.~Kirkpatrick, R.~Pascanu, N.~Rabinowitz, J.~Veness, G.~Desjardins, A.~A.
  Rusu, K.~Milan, J.~Quan, T.~Ramalho, A.~Grabska-Barwinska, D.~Hassabis,
  C.~Clopath, D.~Kumaran, R.~Hadsell, Overcoming catastrophic forgetting in
  neural networks, cite arxiv:1612.00796 (2016).

\bibitem{datastreamevaluation}
J.~Gama, R.~Sebasti{\~{a}}o, P.~P. Rodrigues,
  \href{https://doi.org/10.1007/s10994-012-5320-9}{On evaluating stream
  learning algorithms}, Machine Learning 90~(3) (2013) 317--346 (2013).
\newblock \href {https://doi.org/10.1007/s10994-012-5320-9}
  {\path{doi:10.1007/s10994-012-5320-9}}.
\newline\urlprefix\url{https://doi.org/10.1007/s10994-012-5320-9}

\bibitem{Gamaconceptdrift}
J.~a. Gama, I.~\v{Z}liobaite, A.~Bifet, M.~Pechenizkiy, A.~Bouchachia, A survey
  on concept drift adaptation, ACM Comput. Surv. 46~(4) (2014) 44:1--44:37
  (Mar. 2014).

\bibitem{Hinton_AE}
G.~E. Hinton, R.~S. Zemel, Autoencoders, minimum description length and
  helmholtz free energy, in: Proceedings of the 6th International Conference on
  Neural Information Processing Systems, NIPS'93, Morgan Kaufmann Publishers
  Inc., San Francisco, CA, USA, 1993, pp. 3--10 (1993).

\bibitem{Murphy_Machine_Learning}
K.~P. Murphy, Machine Learning: A Probabilistic Perspective, The MIT Press,
  2012 (2012).

\bibitem{Gama2006}
J.~a. Gama, R.~Fernandes, R.~Rocha, Decision trees for mining data streams,
  Intell. Data Anal. 10~(1) (2006) 23--45 (Jan. 2006).

\bibitem{SCN}
D.~Wang, M.~Li, Stochastic configuration networks: Fundamentals and algorithms,
  IEEE transactions on cybernetics 47~(10) (2017) 3466--3479 (2017).

\bibitem{Bengio_Greedy}
Y.~Bengio, P.~Lamblin, D.~Popovici, H.~Larochelle, Greedy layer-wise training
  of deep networks, in: Proceedings of the 19th International Conference on
  Neural Information Processing Systems, NIPS'06, MIT Press, Cambridge, MA,
  USA, 2006, pp. 153--160 (2006).

\bibitem{goodfellow2016deep}
I.~Goodfellow, Y.~Bengio, A.~Courville, Deep learning, MIT press, 2016 (2016).

\bibitem{xavierinitialization}
Y.~Jia, E.~Shelhamer, J.~Donahue, S.~Karayev, J.~Long, R.~Girshick,
  S.~Guadarrama, T.~Darrell, Caffe: Convolutional architecture for fast feature
  embedding, in: Proceedings of the 22nd ACM international conference on
  Multimedia, ACM, 2014, pp. 675--678 (2014).

\bibitem{woolson2007wilcoxon}
R.~Woolson, Wilcoxon signed-rank test, Wiley encyclopedia of clinical trials
  (2007) 1--3 (2007).

\end{thebibliography}

\end{document}